\documentclass[letterpaper]{article} 
\usepackage{aaai23}
\usepackage{times}  
\usepackage{helvet}  
\usepackage{courier}  
\usepackage[hyphens]{url}  
\usepackage{graphicx} 
\usepackage{natbib}  
\usepackage{caption} 
\usepackage{algorithm}
\usepackage{algorithmic}
\usepackage{amsmath}
\usepackage{cleveref}
\usepackage{amsfonts}
\usepackage{amssymb}
\usepackage{bbm}
\usepackage{booktabs}
\usepackage{bbding}
\usepackage{pifont}
\usepackage{wasysym}

\usepackage{newfloat}
\usepackage{listings}
\DeclareCaptionStyle{ruled}{labelfont=normalfont,labelsep=colon,strut=off} 
\floatstyle{ruled}
\newfloat{listing}{tb}{lst}{}
\floatname{listing}{Listing}
\usepackage{subcaption}

\title{AutoCost: Evolving Intrinsic Cost for Zero-violation Reinforcement Learning}

\author{
    Tairan He,
    Weiye Zhao,
    Changliu Liu
}
\affiliations{
    Robotics Institute, Carnegie Mellon University\\

    \{tairanh, weiyezha, cliu6\}@andrew.cmu.edu 
}

\usepackage{bibentry}

\begin{document}
\maketitle
\begin{abstract}
Safety is a critical hurdle that limits the application of deep reinforcement learning (RL) to real-world control tasks. To this end, constrained reinforcement learning leverages cost functions to improve safety in constrained Markov decision processes. However, such constrained RL methods fail to achieve zero violation even when the cost limit is zero. This paper analyzes the reason for such failure, which suggests that a proper cost function plays an important role in constrained RL. Inspired by the analysis, we propose AutoCost, a simple yet effective framework that automatically searches for cost functions that help constrained RL to achieve zero-violation performance. We validate the proposed method and the searched cost function on the safe RL benchmark Safety Gym. 
We compare the performance of augmented agents that use our cost function to provide additive intrinsic costs with baseline agents that use the same policy learners but with only extrinsic costs. Results show that the converged policies with intrinsic costs in all environments achieve zero constraint violation and comparable performance with baselines.
\end{abstract}

\section{Introduction}
Reinforcement learning (RL) has achieved remarkable progress in board games~\cite{mnih2015human}, card games~\citep{brown2018superhuman} and video games~\citep{vinyals2019grandmaster}. Despite its impressive success so far, the lack of safety guarantees limits the application of RL to real-world physical tasks like robotics. This is particularly concerning for safety-critical scenarios such as robot-human collaboration or healthcare where unsafe controls may lead to fatal consequences. 

\begin{figure}[t]
    \begin{subfigure}[t]{0.23\textwidth}
        \centering
        {\includegraphics[width=\textwidth]{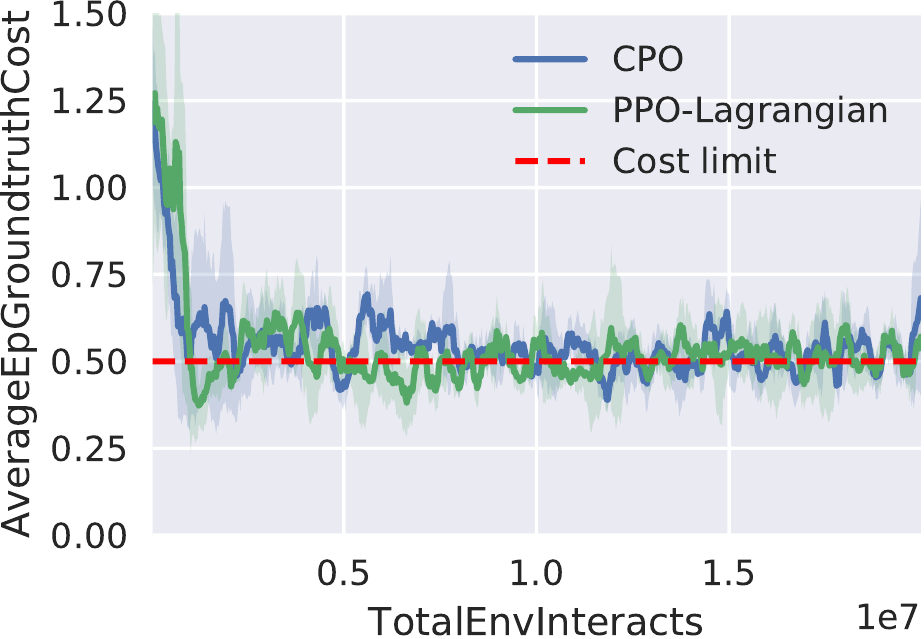}}
        \caption{Cost limit = 0.5}
        \label{fig:intro_limit_0.5}
    \end{subfigure}
    \begin{subfigure}[t]{0.23\textwidth}
        \centering
        {\includegraphics[width=\textwidth]{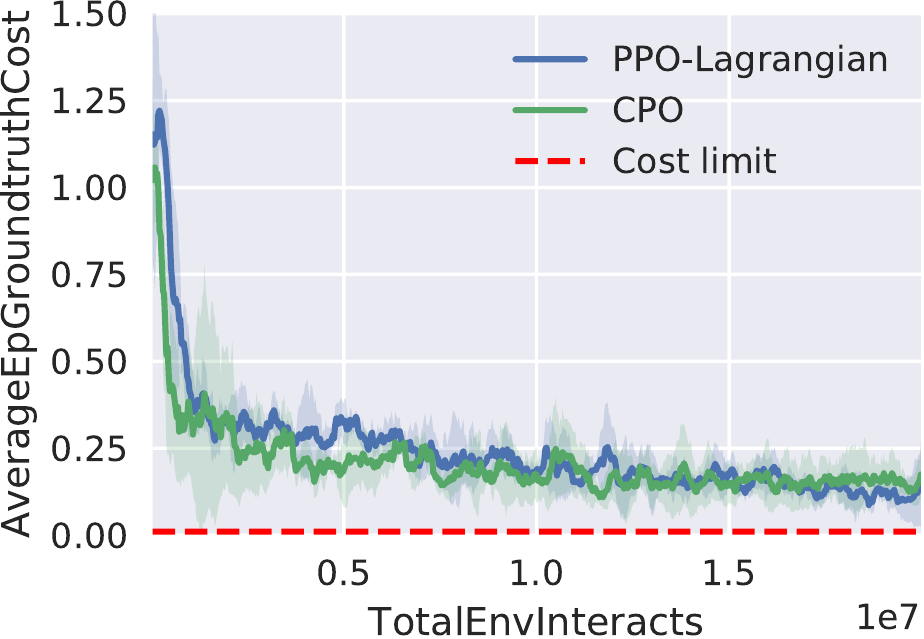}}
        \caption{Cost limit = 0}
        \label{fig:intro_limit_0}
    \end{subfigure}
    \vspace{-5pt}
    \caption{Average episodic extrinsic cost of safe RL baselines with different limit. CPO and PPO-Lagrangian both fail to achieve zero violation even with a zero cost limit}
    \label{fig:intro_limit}
    \vspace{-15pt}
\end{figure}

Many safe RL methods leverage cost functions defined in constrained Markov decision process (CMDP)~\cite{altman1999cmdp} to formulate safety. Recent approaches usually adopt \textit{indicator cost functions} where a positive signal deems a state as unsafe and zero deems a state safe. However, under such a design of cost functions, state-of-the-art safe RL methods still fail to achieve zero violation even with sufficiently enough interactions with the environment~\citep{ma2021learn}. For example, \Cref{fig:intro_limit} illustrates the average episodic extrinsic cost (i.e., constraint violations in CMDP) of two safe RL baselines~\citep{achiam2017cpo,chow2017lagrangian} in the commonly used safe RL benchmark Safety Gym. 

Note that though the average extrinsic cost reduces to 0.5 when the cost limit is set as 0.5, both safe RL baselines with zero cost limit fail to obtain a zero-violation policy even after convergence. Nevertheless, in safety-critical applications, reducing the number of safety violations may not be sufficient since one violation may lead to unacceptable catastrophes. How to eliminate all violations and achieve zero violation remains a challenge for RL.

To break this barrier, in this paper, we conduct an empirical analysis on why safe RL methods fail to achieve zero violation. We find that the reason is closely related to the design of cost functions, where a proper cost function could drive the converged safe RL policies to achieve zero violation. However, designing cost functions has the same issues as designing reward functions~\citep{hong2018diversity,faust2019evolvingreward}, which requires extensive human efforts. To this end, instead of handcrafting a cost function with expert knowledge, we propose AutoCost which automatically searches for proper cost functions with the aim of obtaining a zero-violation policy, relieving researchers from such tedious work. AutoCost formulates the problem as a bi-level optimization where we try to find the best intrinsic cost function, which, to the most extent, helps train a zero-violation RL agent. Specifically, AutoCost utilizes an evolutionary strategy to search the parameters of
cost function. 

We apply the evolutionary intrinsic cost search to an environment in Safety Gym~\citep{ray2019safetygym}, over two safe RL algorithms: Constrained Policy Optimization (CPO)~\citep{achiam2017cpo} and Lagrangian Proximal Policy Optimization (PPO-Lagrangian)~\cite{chow2017lagrangian}. We optimize the parameters of intrinsic cost functions over two objectives: \textit{cost rate} and \textit{reward performance}. The results show that augmented agents that use the searched cost function to provide additive intrinsic costs achieve zero violation under different settings including (i) different control tasks; (ii) different robots; (iii) different types of obstacles. We further analyze why the searched intrinsic cost function boosts the safety of RL agents, revealing that some vital properties of cost functions like local awareness and dense signals are beneficial to safe RL methods. The main contribution of this paper are listed as follows:
\begin{itemize}
    \item We find that different safe RL algorithms react differently to cost functions. We analyze the failure of current safe RL algorithms and point out that a proper cost function is the key to zero-violation RL policies.
    \item To the best of our knowledge, we are the first to propose automated searching for cost function to achieve zero-violation safety performance. 
    \item We evaluate the searched cost function with different safe RL algorithms on state-of-the-art safe RL benchmarks. Results show that the converged policies with intrinsic costs in all environments achieve zero constraint violation and comparable performance with baselines.
\end{itemize}

\section{Related Work}
\subsection{Safe RL} The safe RL algorithms can be divided into three categories.
\begin{itemize}
    \item \textbf{Lagrangian method:} Lagrangian multiplier~\citep{boyd2004convex} provides a regularization term to penalize objectives with constraint violations. Previous work~\citep{chow2017lagrangian} derives a formula to compute the gradient of the Lagrangian function in CMDP. Some safe RL works~\citep{liang2018accelerated,tessler2018reward} solve a primal-dual optimization problem to satisfy constraints. 
    Another work~\citep{stooke2020responsive} leverages derivatives of the constraint function and PID Lagrangian methods to reduce violations during agent learning.
    More recent work~\citep{chen2021primal} gives an analysis on the convergence rate for sampling-based primal-dual Lagrangian optimization. 
    \item \textbf{Direct policy optimization methods:} Though Lagrangian methods are easy to transfer to the setting of safe RL, the resulted policy was only safe asymptotically and lacks safety guarantees during each training iteration~\citep{chow2018lyapunov}. Therefore, many works propose to derive surrogate algorithms with direct policy optimization. One representative method is constrained policy optization~\citep{achiam2017cpo} which leverages the trust region of constraints for policy updates. Another work~\citep{el2016convexsynthesis} proposes a conservative stepwise surrogate constraint and leverages linear programming to generate feasible safe policies. FOCOPS~\citep{zhang2020first} proposes a convex approximation to replace surrogate functions with first-order Taylor expansion. 
    \item \textbf{Safeguard-based methods:} Another line of safe RL works add a safeguard upon the policy network, projecting unsafe actions to safe actions. Some works~\citep{srinivasan2020learningtobesafe,bharadhwaj2020conservative} leverage a safety critic to detect unsafe conditions and use such a critic to choose the safe action. Safety layer~\citep{dalal2018safe} solves a QP problem to safeguard unsafe actions with learned dynamics, but it is limited to linear cost functions. ISSA~\citep{zhao2021issa} proposes an efficient sampling algorithm for safe actions and guarantees zero violation during agent learning, but it also requires a perfect dynamics model for a one-step prediction. ShieldNN~\citep{ferlez2020shieldnn} designs a safety filter neural network with safety guarantees specifically for the kinematic bicycle model (KBM) \cite{kong2015kinematic}, which limits the generalization ability to other dynamics. 
\end{itemize}
Besides different methodology, the types of safety considered by these safe RL algorithms are also different.
Most Lagrangian and direct policy optimization methods consider safety as the cost expectation of trajectories being lower than a cost limit (i.e., \textit{safety in expectation}), while safeguard-based methods consider \textit{state-wise safety}. 
However, safeguard-based methods usually require either the dynamics model or prior knowledge as discussed above, which is unrealistic in practice.
Our work aims to boost general safe RL algorithms (i.e., the first two categories) to achieve safety in expectation but with a zero cost limit. Note that strictly satisfying a zero cost limit in expectation is actually equivalent to state-wise safety. So our framework is a vital step that bridges \textit{safety in expectation} and \textit{state-wise safety}.

\subsection{Certificate of Safety}
Besides the cost function of CMDP used in safe RL, there are many different certificates~\citep{wei2019safe} to measure safety in the safe control community including (i) potential function; (ii) safety index; (iii) control barrier function. Representative methods include potential field method~\citep{khatib1986real}, sliding mode algorithm~\citep{gracia2013reactive}, barrier function method~\citep{ames2014control}, and safe set algorithm~\citep{liu2014control}. However, designing such safety certificate functions is difficult and requires great human efforts to find appropriate parameters and functional forms~\citep{dawson2022survey,zhao2021issa}. 
Some automated synthesis techniques for safety certificates are either computationally expensive~\citep{giesl2015review} or limited to specific dynamics~\citep{ahmadi2016some}.
Recently, many works~\citep{chang2019neural,qin2021learning,dawson2022safe} leverage neural networks to learn such safety certificates. But such methods require prior signals (e.g., safe and unsafe regions) for supervised learning, whereas our framework searches for proper intrinsic cost functions automatically with the goal of zero violation.

\subsection{Automating RL} How to automate RL has been an active topic recently. 
Some works leverage techniques from the AutoML community to automate the process of hyper-parameters searching~\citep{paul2019fast,zahavy2020self,xu2020meta} and architectures searching~\citep{runge2018rna,franke2020sample} for RL.
Besides, designing a reward function that encourages the desired behaviors while still being learnable for RL is difficult~\citep{dewey2014reinforcement}.
To this end, AutoRL~\cite{faust2019evolvingreward} proposes to automatically search for better reward functions in an evolutionary manner. 
Another work~\citep{veeriah2019discovery} proposes to discover more helpful value/reward functions using meta gradients.
LIRPG~\citep{zheng2018intrinsic} proposes an algorithm for learning intrinsic rewards for RL. To the best of our knowledge, we are the first to automate the process of designing cost functions for zero-violation safe RL.

\section{Problem Formulation and Background}
\subsection{Constrained Markov Decision Process}
We consider constrained Markov decision process (CMDP) defined as the tuple $(\mathcal{S}, \mathcal{A}, P, \mathcal{R}, \mathcal{C}, \rho_0, \gamma)$ with following components: states $s \in \mathcal{S}$, actions $a \in \mathcal{A}$, P is the distribution of state transition where $P = P(s_{t+1} | s_t, a_t)$, $\rho_0: \mathcal{S} \rightarrow [0, 1]$ is the initial state distribution, $\gamma \in [0, 1]$ is the discounted factor. The agent holds
its policy $\pi(a|s): \mathcal{S} \times \mathcal{A} \rightarrow [0, 1]$ to make decisions and receive rewards defined as $r: \mathcal{S} \rightarrow \mathbb{R}$ and cost defined as $c: \mathcal{S} \rightarrow \mathbb{R}$. To distinguish between the cost function defined in the CMDP and other additive cost functions, we denote the original cost function in the CMDP as the \textit{extrinsic cost} as $c^{ex}: \mathcal{S} \rightarrow \mathbb{R}$, and we denote \textit{intrinsic cost} parameterized by $\theta$ as $c^{in}_\theta: \mathcal{S} \rightarrow \mathbb{R}$. 

In RL, we aim to select a policy $\pi$ which maximizes the discounted cumulative rewards $\mathcal{R}^\pi := \mathbb{E}_{\tau \sim \pi}[\sum_{t=0}^{\infty}\gamma^t r_t]$. Meanwhile, the extrinsic cost function $c^{ex}$ in CMDP defines the cost return as $J^\pi_{c^{ex}} := \mathbb{E}_{\tau \sim \pi}[\sum_{t=0}^{\infty}\gamma^t c^{ex}_t]$. Based on these two definitions, the feasible policies $\Pi_{c^{ex}}$ and optimal policies $\pi^*$ in CMDP are defined as:
\begin{equation}
\Pi_{c^{ex}} := \{\pi \in \Pi: J^\pi_{c^{ex}} \leq d \}, \pi^* = \mathop{\arg\max}_{\pi \in \Pi_{c^{ex}}} \mathcal{R}^\pi
\label{eq:feasible policy and optimal policy}
\end{equation}
The goal of constrained RL algorithms is:
\begin{equation}
\max_{\pi \in \Pi} \mathcal{R}^{\pi} \quad \text{ s.t. } \  J^\pi_{c^{ex}} \leq d
\label{eq:goal of contained RL}
\end{equation}
\subsection{Proximal Policy Optimization}
Proximal policy optimization (PPO)~\citep{schulman2017proximal} is an on-policy policy gradient algorithm. PPO proposes a surrogate objective that attains the data efficiency and reliable performance of TRPO~\citep{schulman2015trpo} while using only first-order optimization:
\begin{equation}
\begin{split}
\pi_{k+1} = & \mathop{\arg\min}_{\pi \in \Pi} \mathbb{E}_{\tau\sim\pi_k} \min \Big(\frac{\pi(a|s)}{\pi_k(a|s)}A^{\pi_k}_r(s,a), \\
&\quad \text{clip}\big(\frac{\pi(a|s)}{\pi_k(a|s)}, 1-\epsilon, 1+\epsilon \big)A^{\pi_k}_r(s,a) \Big)~,
\end{split}
\label{eq:ppo}
\end{equation}
where $A^{\pi_k}_r$ is the advantage function of reward with respect to $\pi_k$ and $\epsilon$ is a (small) hyper-parameter determining how far away the new policy $\pi_{k+1}$ is allowed to go from the old $\pi_{k}$.

\subsection{Lagrangian Methods} Lagrangian methods~\citep{boyd2004convex} use adaptive coefficients to enforce constraints in optimization with $f(\omega)$ the objective and $g(\omega) \leq 0$ the constraint:
\begin{equation}
\max_{\omega} \min_{\lambda \geq 0} \mathcal{L}(\omega, \lambda) = f(\omega) -\lambda g(\omega)
\label{eq:simple lagrangian}
\end{equation}
Similarly, Lagrangian safe  RL~\citep{chow2017lagrangian} adopt such an iterative learning scheme to solve \Cref{eq:goal of contained RL} as:
\begin{equation}
\max_{\pi} \min_{\lambda \geq 0} \mathcal{L}(\pi, \lambda) = \mathcal{R}^{\pi} -\lambda (J^\pi_{c^{ex}} - d)
\label{eq:lagrangian RL}
\end{equation}
In the implementation, Lagrangian safe  RL methods iteratively take gradient ascent steps with respect to $\pi$ and gradient descent with respect to $\lambda$.

\subsection{Constrained Policy Optimization}
CPO~\citep{achiam2017cpo} analytically solves trust region optimization problems at each policy update to enforce constraints throughout training:
\begin{equation}
\begin{split}
\pi_{k+1} & = \mathop{\arg\max}_{\pi \in \Pi} \mathbb{E}_{s\sim\pi_k, a\sim\pi}[A^{\pi_k}_r(s,a)]\\
\text{s.t.} \quad & J^{\pi_k}_{c^{ex}} + \frac{1}{1-\gamma} \mathbb{E}_{s\sim\pi_k, a\sim\pi}[A^{\pi_k}_{c^{ex}}(s,a)] \leq d \\
& \hat{D}(\pi || \pi_k) \leq \delta~,
\end{split}
\label{eq:cpo updates}
\end{equation}
where $A^{\pi_k}_r$ and $A^{\pi_k}_{c^{ex}}$ are advantage functions of reward and extrinsic cost functions with respect to $\pi_k$.

\section{Empirical Observation and Analysis}
In this section, we begin with empirical observations that motivate the core idea of our method to design cost functions in order to achieve zero violation. Generally, we aim to investigate why safe RL methods fail to achieve zero violation as shown in \Cref{fig:intro_limit}. We come up with two hypotheses and verify them in the following subsections.
\subsection{More Conservative Cost Limit}
\begin{figure}[htbp]
    \begin{subfigure}[t]{0.23\textwidth}
        \centering
        {\includegraphics[width=\textwidth]{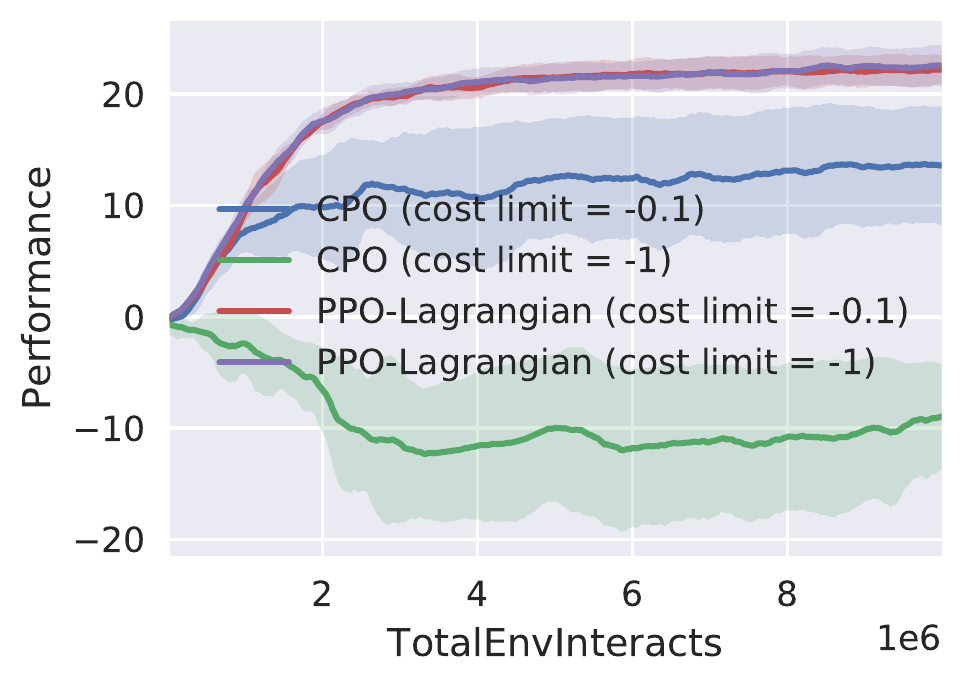}}
        \caption{Performance}
        \label{fig:empirical_cost_limit_n0.1-n1.0_performance}
    \end{subfigure}
    \begin{subfigure}[t]{0.23\textwidth}
        \centering
        {\includegraphics[width=\textwidth]{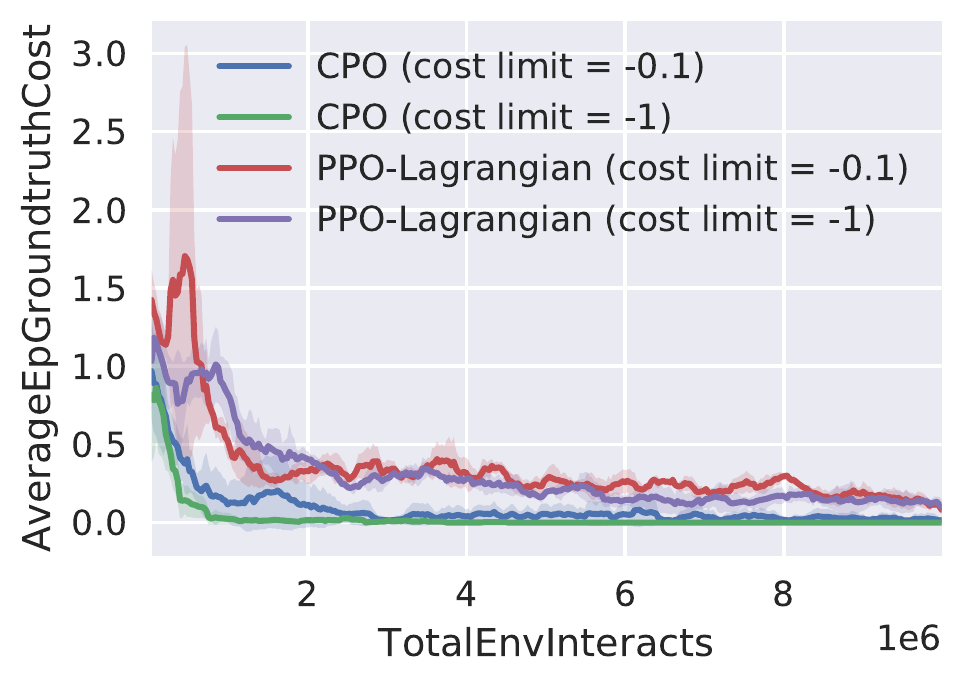}}
        \caption{Extrinsic cost}
        \label{fig:empirical_cost_limit_n0.1-n1.0_cost}
    \end{subfigure}
    \vspace{-5pt}
    \caption{Average episodic reward and cost of safe RL baselines with negative cost limit.}
    \label{fig:empirical_cost_limit_n0.1-n1.0}
    \vspace{-15pt}
\end{figure}

Inspired by \Cref{fig:intro_limit}, we find that safe RL does not strictly meet the requirements of the cost limit and oscillates around the threshold. To further eliminate violations, we attempt a more conservative cost limit (i.e., a negative cost limit). The results are shown in \Cref{fig:empirical_cost_limit_n0.1-n1.0}, where CPO achieves near zero violation with the cost limit of $-0.1$ and achieves zero violation with the cost limit of $-1.0$, but with significant performance drops. On the contrary, negative cost limits do not affect PPO-Lagrangian differently compared with the zero cost limit. The reason behind the difference is that a negative cost limit makes CPO update policies in the most conservative way (i.e., only considering constraints). As for Lagrangian methods, a negative cost limit just applies the relatively same penalty on objectives at each policy update. 
We conclude that tuning the cost limit is possible to help CPO to achieve zero violation (with significant performance drops) but can not help Lagrangian methods. To find a more general method to drive safe RL agents to achieve zero violation, we turn to the design of cost functions.

\subsection{Denser Cost Function}
\begin{figure}[htbp]
    \vspace{-5pt}
    \begin{subfigure}[t]{0.23\textwidth}
        \centering
        {\includegraphics[width=\textwidth]{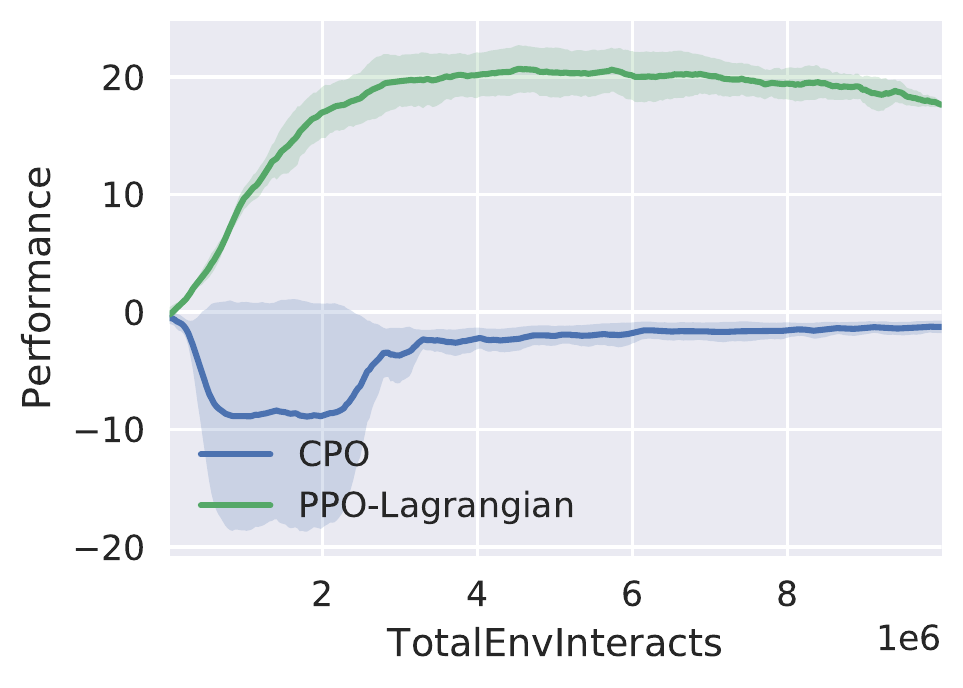}}
        \caption{Performance}
        \label{fig:empirical_dense_cost_performance}
    \end{subfigure}
    \begin{subfigure}[t]{0.23\textwidth}
        \centering
        {\includegraphics[width=\textwidth]{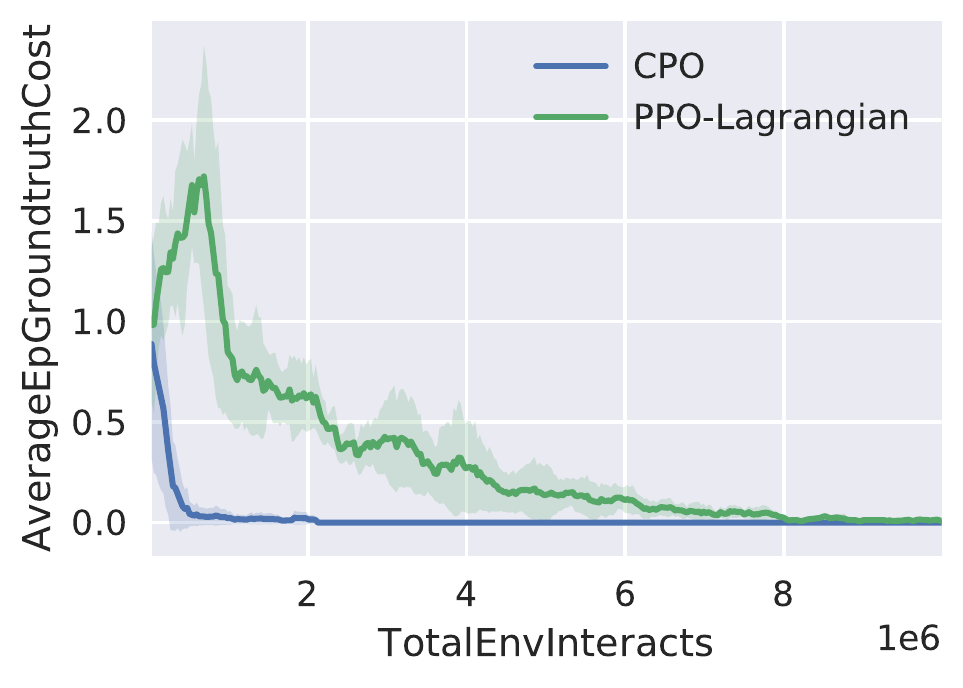}}
        \caption{Extrinsic cost}
        \label{fig:empirical_dense_cost_cost}
    \end{subfigure}
    \vspace{-5pt}
    \caption{Average episodic reward and cost of safe RL baselines with a dense cost function.}
    \label{fig:empirical_dense_cost}
    \vspace{-15pt}
\end{figure}

The sparsity (costs only obtained on constraint violation) of indicator cost functions has the same issue of sparse reward functions~\citep{andrychowicz2017hindsight}. Similarly, we apply a commonly used dense reward design to cost functions:
\begin{equation}
c(s_t) = \max(d(s_{t-1}) - d(s_t), 0)~,
\label{eq:empirical_dense_cost}
\end{equation}
where $d(\cdot)$ denotes the distance from RL agent to the closet constraint. Note that this dense cost function is very conservative because it assigns positive costs to any step making the RL agent closer to constraints. We train CPO and PPO-Lagrangian under the denser cost function. The results are shown in \Cref{fig:empirical_dense_cost}, where we find that the dense cost function is detrimental to RL performance of CPO though it helps CPO quickly converges to a zero-violation policy. On the other hand, PPO-Lagrangian converges to a near zero-violation policy in the end with satisfying reward performance. We point out that safe RL algorithms react differently under the same cost function due to different algorithmic designs. One cost function may be suitable for a specific algorithm but may be detrimental to another algorithm. How to find a more generalizable cost function for different safe RL algorithms remains a challenge. We also try more hand-designed cost functions (details are given in \Cref{appendix:manually_designed}), but none is good enough to achieve both zero violation and satisfying reward performance, indicating the difficulty of manually designing cost functions.

\section{Automating Cost Function Design}
\subsection{Bi-level Optimization Problem} In this paper, we aim to find the intrinsic cost function $c^{in}_{\theta}$ (parameterized by $\theta$) such that the agent $\pi_\omega$ (parameterized by $\omega$) can achieve zero violation by maximizing cumulative discounted rewards $\mathcal{R}^\pi$ under the constraint of cost function $c^{ex+in}_{\theta} = c^{ex} +  c^{in}_{\theta}$. Formally, our optimization goal is:
\begin{equation}
\begin{split} 
\min_{\theta} & \quad  J^{\pi^*}_{c^{ex}} \\
 \text{s.t.} & \quad \pi^* = \mathop{\arg\max}_{\pi\in\Pi_{c^{ex+in}_{\theta}}}\mathcal{R}^\pi.
\end{split} 
\label{eq:nested_optimization}
\end{equation}
Compared with other possible formulations (e.g., considering $J^{\pi^*}_{c^{ex+in}_\theta}$ in the outer loop), the objective in \Cref{eq:nested_optimization} tries to minimize the ground truth constraint violation (i.e., the extrinsic cost objective $J^{\pi^*}_{c^{ex}}$) to achieve zero violation. This formulation enables safe RL to further eliminate constraint violations.
And the corresponding $\pi^*$ is solved with downstream safe RL algorithms within the feasible policy set $\Pi_{c^{ex+in}_\theta}$ defined by $c^{ex+in}_\theta$. 

To find proper cost functions without expert knowledge, we propose AutoCost, a principled solution for resolving the bi-level optimization problem in \Cref{eq:nested_optimization}.
\subsection{AutoCost} 
\begin{figure}[htbp]
    \vspace{-10pt}
    \centering
    \includegraphics[width=0.30\textwidth]{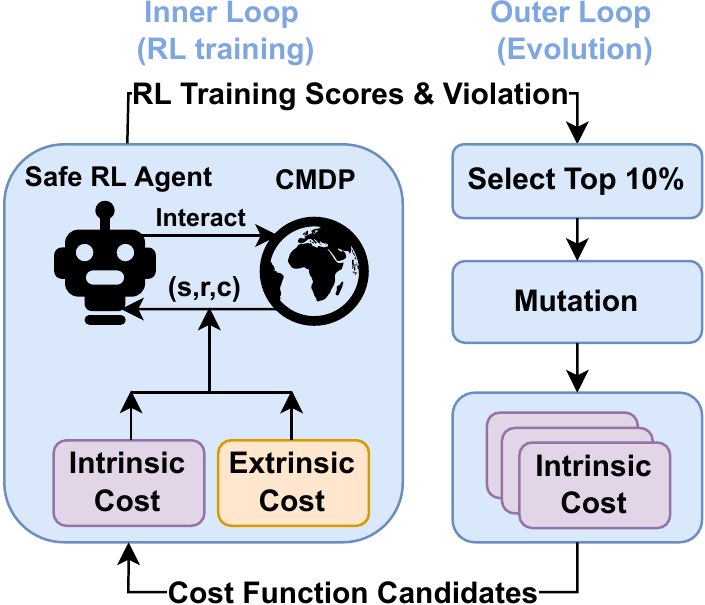}
    \caption{Overview of AutoCost. AutoCost contains an inner loop (left) and an outer loop (right). The inner loop performs an RL training procedure with searched intrinsic cost functions. The outer loop searches intrinsic cost functions using an evolutionary algorithm.}
    \label{fig:overview_autocost}
    \vspace{-10pt}
\end{figure}
\begin{figure*}[t]
    \hfill
    \begin{subfigure}[t]{0.13\textwidth}
        \centering
        {\includegraphics[width=\textwidth]{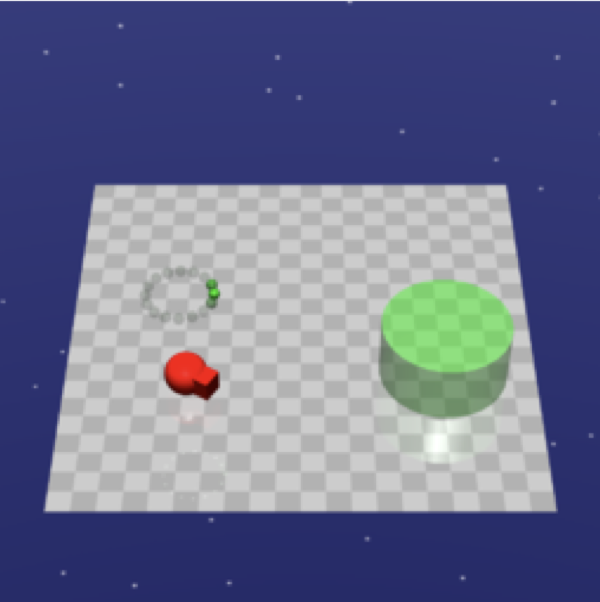}}
        \caption{Goal}
        \label{fig:safety_gym_goal}
    \end{subfigure}
    \hfill
    \begin{subfigure}[t]{0.13\textwidth}
        \centering
        {\includegraphics[width=\textwidth]{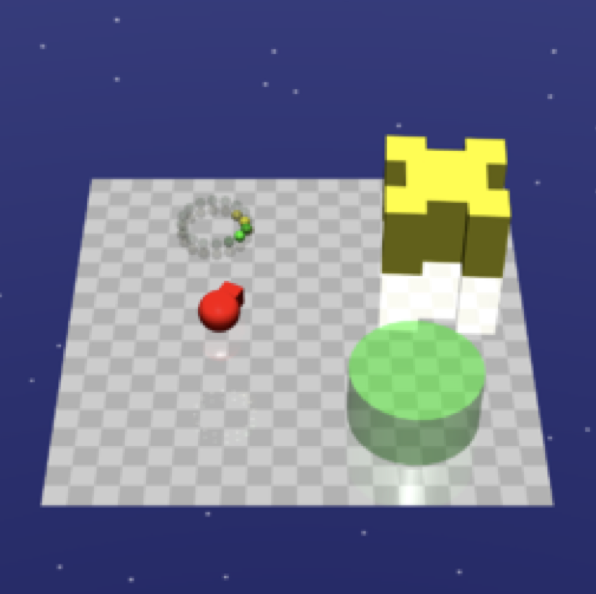}}
        \caption{Push}
        \label{fig:safety_gym_push}
    \end{subfigure}
    \hfill
    \begin{subfigure}[t]{0.13\textwidth}
        \centering
        {\includegraphics[width=\textwidth]{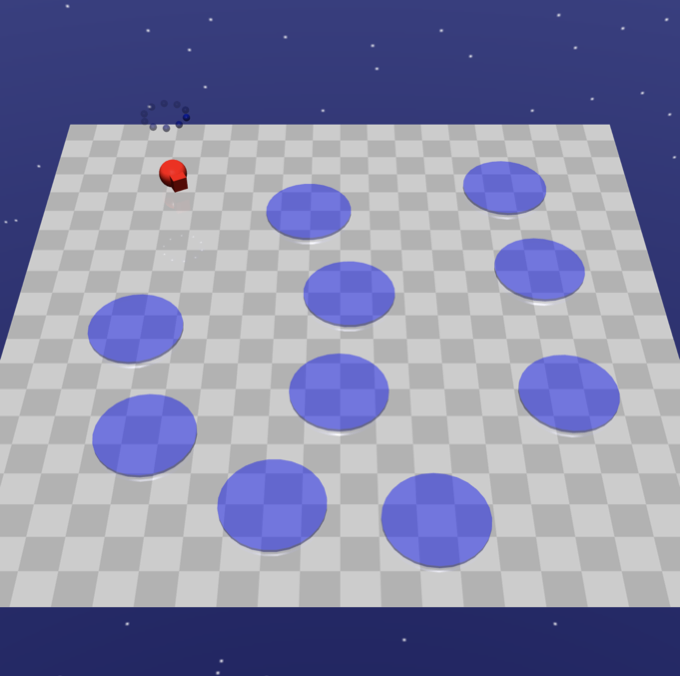}}
        \caption{Hazard}
        \label{fig:safety_gym_hazard}
    \end{subfigure}
    \hfill
    \begin{subfigure}[t]{0.13\textwidth}
        \centering
        {\includegraphics[width=\textwidth]{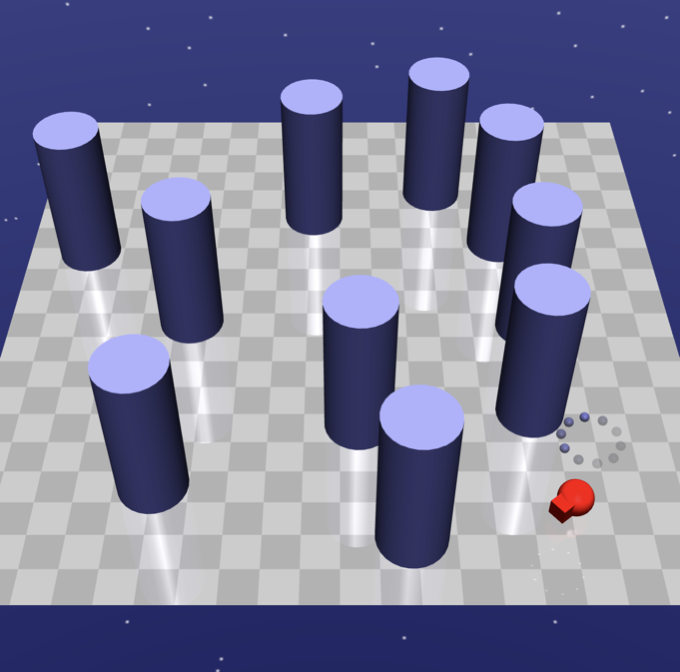}}
        \caption{Pillar}
        \label{fig:safety_gym_pillar}
    \end{subfigure}
    \hfill
    \begin{subfigure}[t]{0.13\textwidth}
        \centering
        {\includegraphics[width=\textwidth]{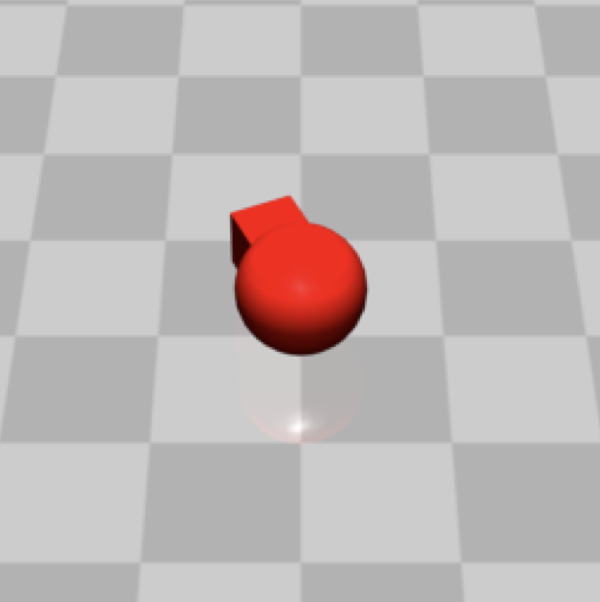}}
        \caption{Point}
        \label{fig:safety_gym_point}
    \end{subfigure}
    \hfill
    \begin{subfigure}[t]{0.13\textwidth}
        \centering
        {\includegraphics[width=\textwidth]{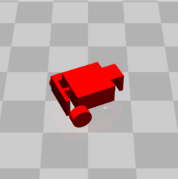}}
        \caption{Car}
        \label{fig:safety_gym_car}
    \end{subfigure}
    \hfill
    \begin{subfigure}[t]{0.13\textwidth}
        \centering
        {\includegraphics[width=\textwidth]{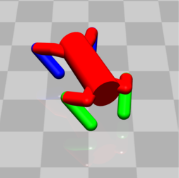}}
        \caption{Doggo}
        \label{fig:safety_gym_doggo}
    \end{subfigure}
\label{fig:safegy_gym_environments}
\vspace{-5pt}
\caption{Different tasks, constraints and robots in Safety Gym.}
\vspace{-15pt}
\end{figure*}
The success of evolution strategies in exploring large, multi-dimensional search space has been proven in many works~\citep{houthooft2018evolvedPG,faust2019evolvingreward,co2020evolvingRL}. Similarly, AutoCost adopts an evolutionary algorithm~\citep{back1993overview} to search for proper cost functions over the parameter space of the intrinsic cost function. The overall pipeline of AutoCost is shown in \Cref{fig:overview_autocost}. Specifically, the evolutionary search of AutoCost contains four parts. 
\begin{enumerate}
    \item \textbf{Initialization}: At the beginning of evolutionary search, randomly sample a population of 50 candidates of intrinsic cost functions $c^{in}_{\theta}$.
    \item \textbf{Evaluation}: At each evolution stage, we train RL agents with the population of intrinsic cost functions.
    \item \textbf{Selection}: After the inner loop of RL training, all candidates are then sorted by the average episodic extrinsic cost of the converged policy. The top-10\% candidates with the lowest constraint violations are selected to generate the population for the next stage. 
    Note that for a population with more than 10\% candidates achieving zero violation, we select the top-10\% candidates with highest reward performance for the next generation.
    \item \textbf{Mutation} After selection, we apply mutation on the top intrinsic cost functions to construct the next generation. We design two types of mutations on parameters $\theta$: (i) Gaussian Noise: $\theta' = \theta + z$, where $z \sim \mathcal{N}(0, I)$; (ii) random scaling: $\theta' = \theta * z$, where $z \sim \text{Uni}[\alpha, \beta]$. 
\end{enumerate}

\section{Experiment}
\subsection{Environment Setup}
We evaluate AutoCost in Safety Gym~\citep{ray2019safetygym}, a widely used benchmark for safe RL algorithms. Safety Gym is built on an advanced physics engine MuJoCo~\citep{todorov2012mujoco}, with various tasks, constraints and robots. We name these environments as \texttt{\{Task\}-\{Constraint Type\}-\{Robot\}}. 
In our experiments, two tasks are considered:
\begin{itemize}
    \item \texttt{Goal}: The robot must navigate inside the green goal area as shown in~\Cref{fig:safety_gym_goal}.
    \item \texttt{Push}: The robot must push a yellow box inside the green goal area as shown in~\Cref{fig:safety_gym_push}.
\end{itemize}
Two different types of constraints are considered:
\begin{itemize}
    \item \texttt{Hazard}: Dangerous (but non-physical) areas as shown in~\Cref{fig:safety_gym_hazard}. The agent is penalized for entering them.
    \item \texttt{Pillar}: Fixed obstacles as shown in ~\Cref{fig:safety_gym_pillar}. The agent is penalized for hitting them.
\end{itemize}
And three robots are considered:
\begin{itemize}
    \item \texttt{Point}: A simple mobile robot with one actuator for turning and another for moving forward/backwards as shown in~\Cref{fig:safety_gym_point}.
    \item \texttt{Car}: A wheeled robot with two independently-driven parallel wheels and a free rolling rear wheel as shown in~\Cref{fig:safety_gym_car}.
    \item \texttt{Doggo}: A quadrupedal robot with bilateral symmetry as shown in~\Cref{fig:safety_gym_doggo}. Each of the four legs has two controls at the hip, for azimuth and elevation relative to the torso, and one in the knee, controlling angle.
\end{itemize}

\subsection{Evolution on Safety Gym}
\begin{figure}[H]
     \centering
     \vspace{-5pt}
     \includegraphics[width=0.32\textwidth]{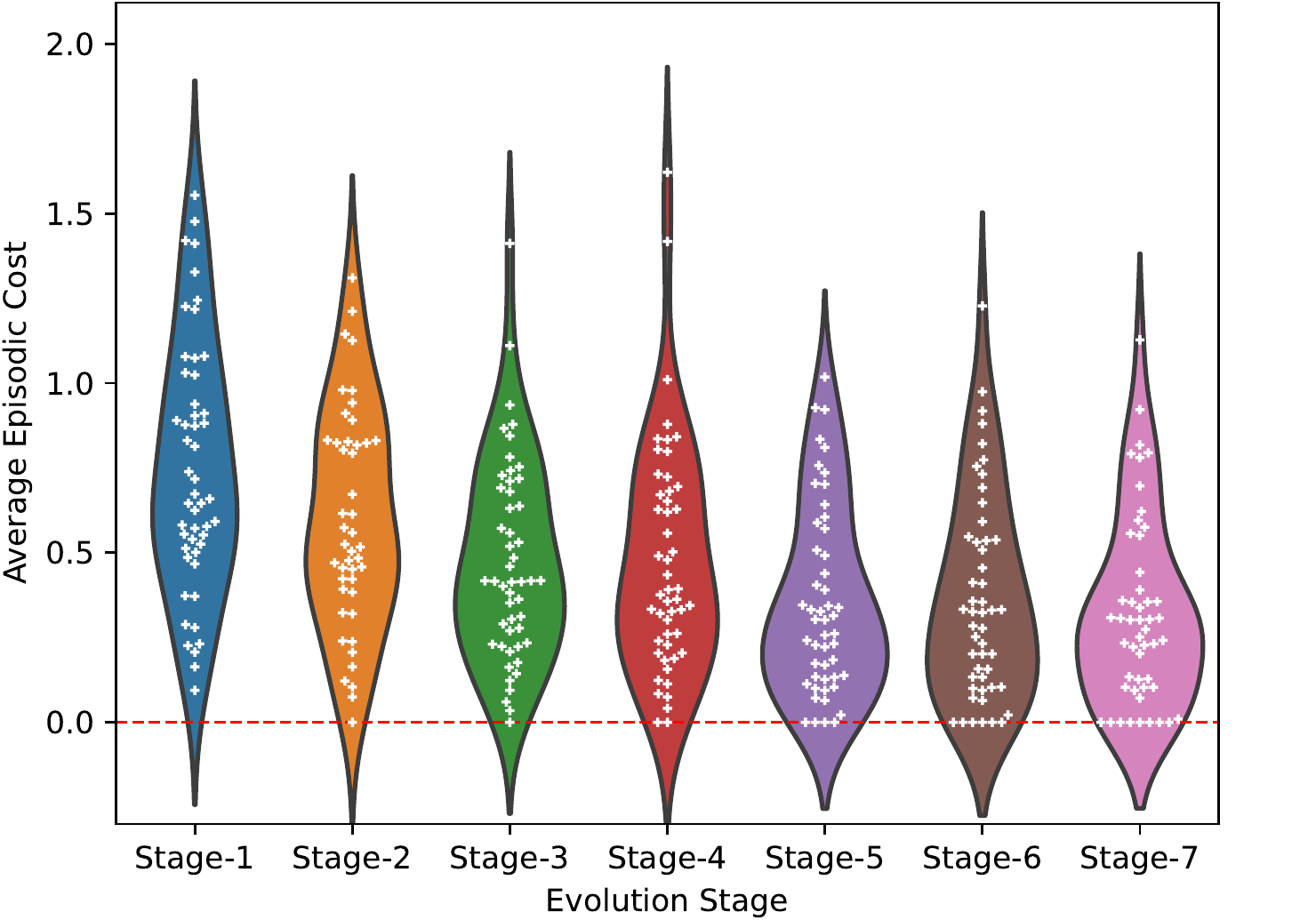}
      \vspace{-8pt}
     \caption{
     Evolution process in the training environment. Every white dot represents a candidate of the intrinsic cost function, and the y-axis shows its corresponding constraint violations after convergence. The red horizontal line indicates zero-violation safety. 
    }
     \label{fig:evolution_process}
     \vspace{-18pt}
\end{figure}

\paragraph{Experiment Setting} We apply AutoCost to environment \texttt{Goal-Hazard-Point}. To ensure the generalizability of searched intrinsic cost functions to different safe RL algorithms, we train two RL agents (CPO and PPO-Lagrangian) for each intrinsic cost function candidate, and we take the average episodic extrinsic costs of CPO and PPO-Lagrangian as the evaluation metric. As for the parameter space of the intrinsic cost function, we use a simple multi-layer perception (MLP) neural network with one hidden layer with four neurons, which results in 41 parameters in total including weights and bias. To enforce the intrinsic cost function to be non-negative, we choose the Sigmoid function as the activation function. To further reduce the size of the parameter space, we only include sensors related to constraints (e.g., hazard/pillar lidar) into the input for the intrinsic cost function. Details about input and parameter space of intrinsic cost functions are given in \Cref{appendix: Intrinsic Cost Settings}.
\paragraph{Evolutionary Process} The evolution process on the training environment \texttt{Goal-Hazard-Point} is shown in \Cref{fig:evolution_process}. We can observe a clear decreasing trend in the average extrinsic cost of populations as evolution continues, indicating the effectiveness of the evolutionary search. 
More importantly, the decreasing trend in constraint violation is especially significant in terms of the best intrinsic cost functions in each population. To be more specific, in the population of stage 1, there is no intrinsic cost achieving zero violation. As the population evolves, more and more intrinsic cost functions drive the safe RL algorithms to obtain a zero-violation policy after convergence. Note that the training of each intrinsic cost candidate is parallelizable. More details about computing time can be found in \Cref{appendix: Computing Infrastructure}.
\begin{figure*}[t]
    \hfill
    \centering
    \begin{subfigure}[t]{0.23\textwidth}
    \begin{subfigure}[t]{1.00\textwidth}
        \raisebox{-\height}{\includegraphics[width=\textwidth]{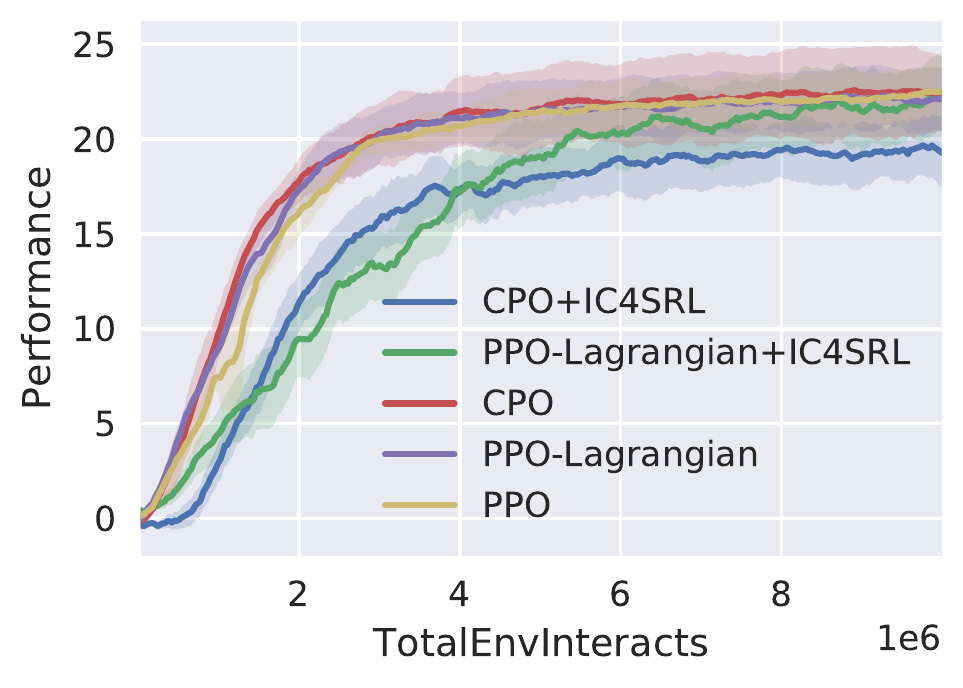}}
        \label{fig:goal-hazard-point-Performance}
    \end{subfigure}
    \hfill
    \begin{subfigure}[t]{1.00\textwidth}
        \raisebox{-\height}{\includegraphics[width=\textwidth]{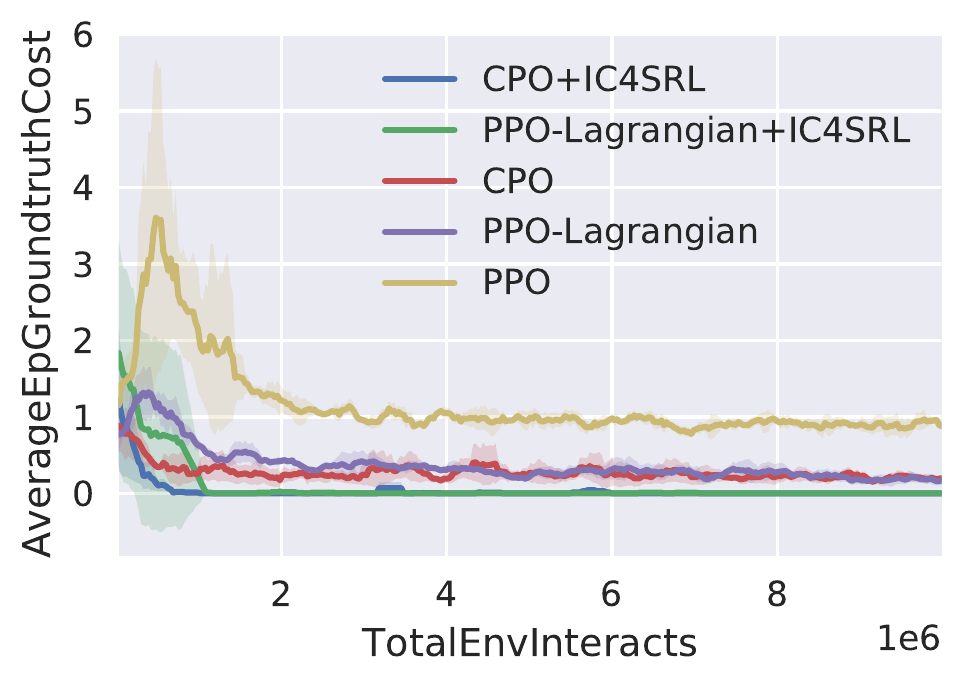}}
        \label{fig:goal-hazard-point-Cost}
    \end{subfigure}
    \caption{Goal-Hazard-Point}
    \label{Goal-Hazard-Point}
    \end{subfigure}
    \begin{subfigure}[t]{0.23\textwidth}
    \begin{subfigure}[t]{1.00\textwidth}
        \raisebox{-\height}{\includegraphics[width=\textwidth]{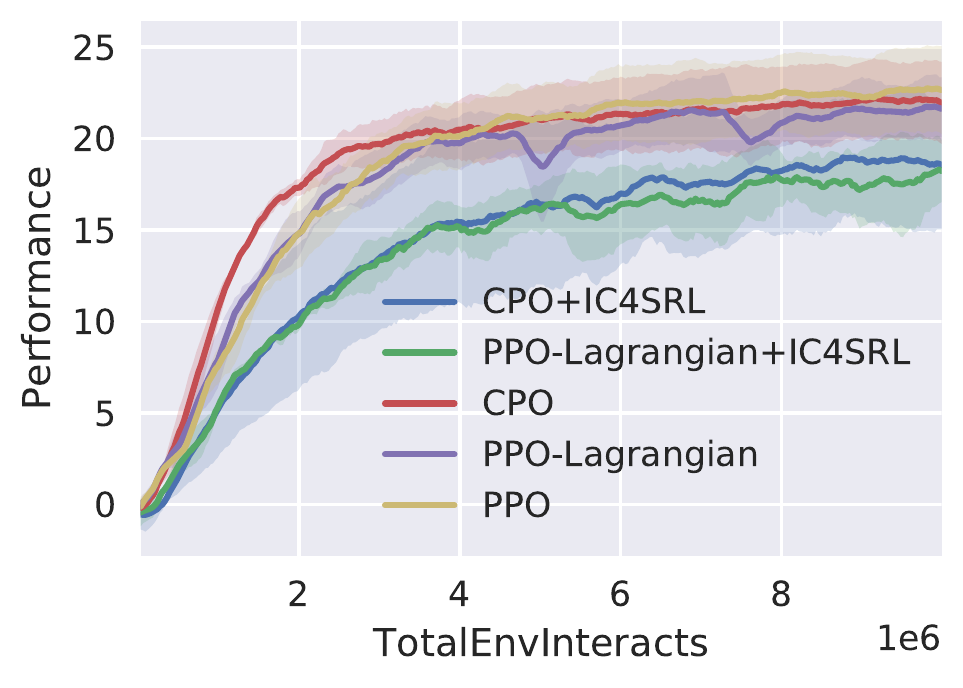}}
        \label{fig:goal-pillar-point-Performance}
    \end{subfigure}
    \hfill
    \begin{subfigure}[t]{1.00\textwidth}
        \raisebox{-\height}{\includegraphics[width=\textwidth]{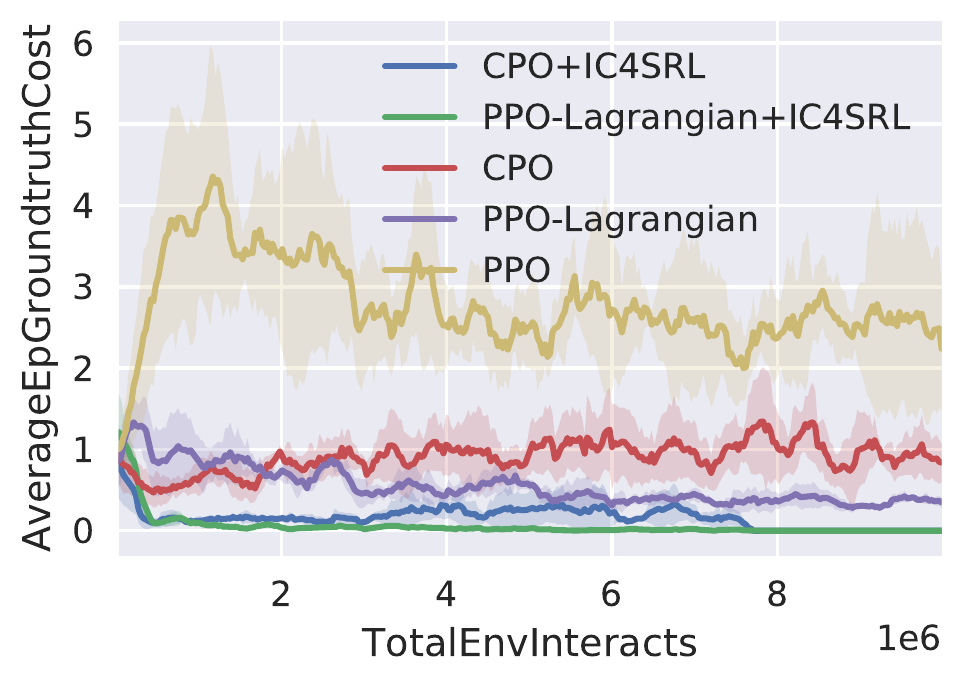}}
        \label{fig:goal-pillar-point-Cost}
    \end{subfigure}
    \caption{Goal-Pillar-Point}
    \label{Goal-Pillar-Point}
    \end{subfigure}
    \begin{subfigure}[t]{0.23\textwidth}
    \begin{subfigure}[t]{1.00\textwidth}
        \raisebox{-\height}{\includegraphics[width=\textwidth]{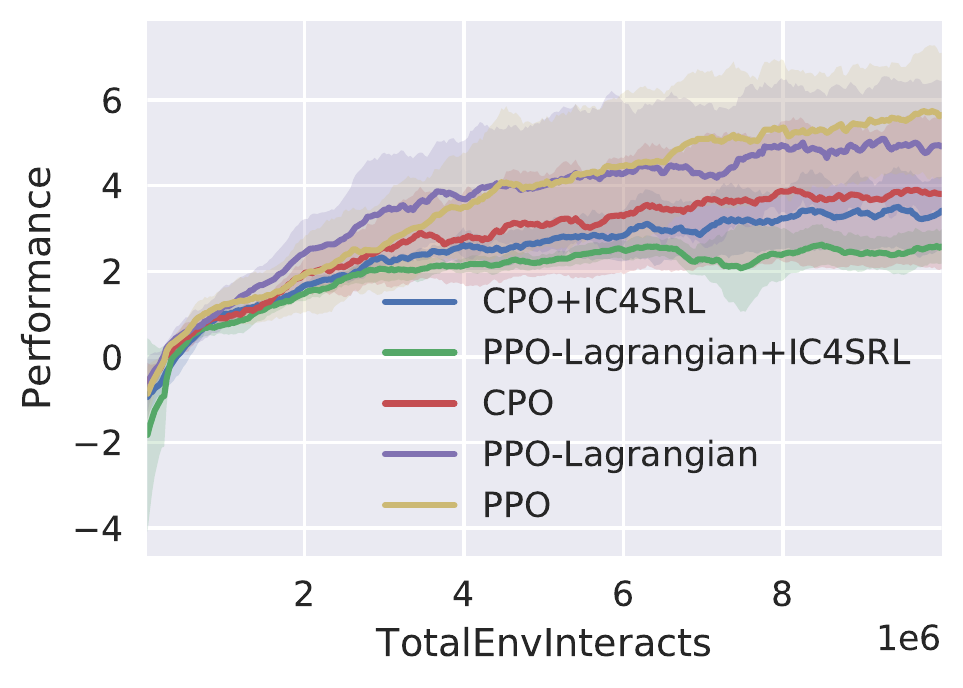}}
        \label{fig:push-hazard-point-Performance}
    \end{subfigure}
    \hfill
    \begin{subfigure}[t]{1.00\textwidth}
        \raisebox{-\height}{\includegraphics[width=\textwidth]{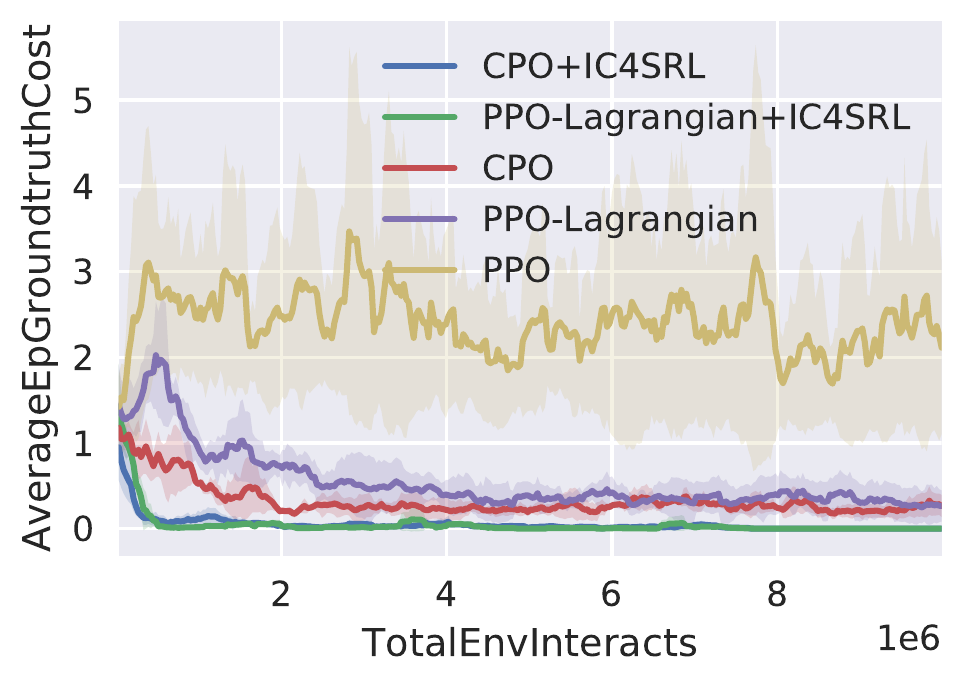}}
        \label{fig:push-hazard-point-Cost}
    \end{subfigure}
    \caption{Push-Hazard-Point}
    \label{Push-Hazard-Point}
    \end{subfigure}
    \begin{subfigure}[t]{0.23\textwidth}
    \begin{subfigure}[t]{1.00\textwidth}
        \raisebox{-\height}{\includegraphics[width=\textwidth]{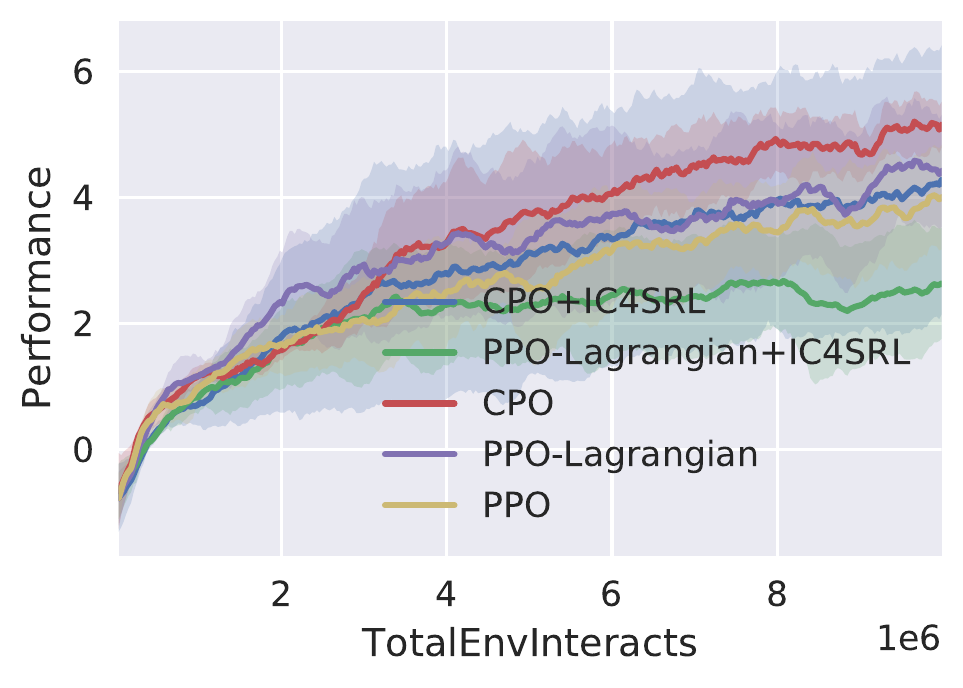}}
        \label{fig:push-pillar-point-Performance}
    \end{subfigure}
    \hfill
    \begin{subfigure}[t]{1.00\textwidth}
        \raisebox{-\height}{\includegraphics[width=\textwidth]{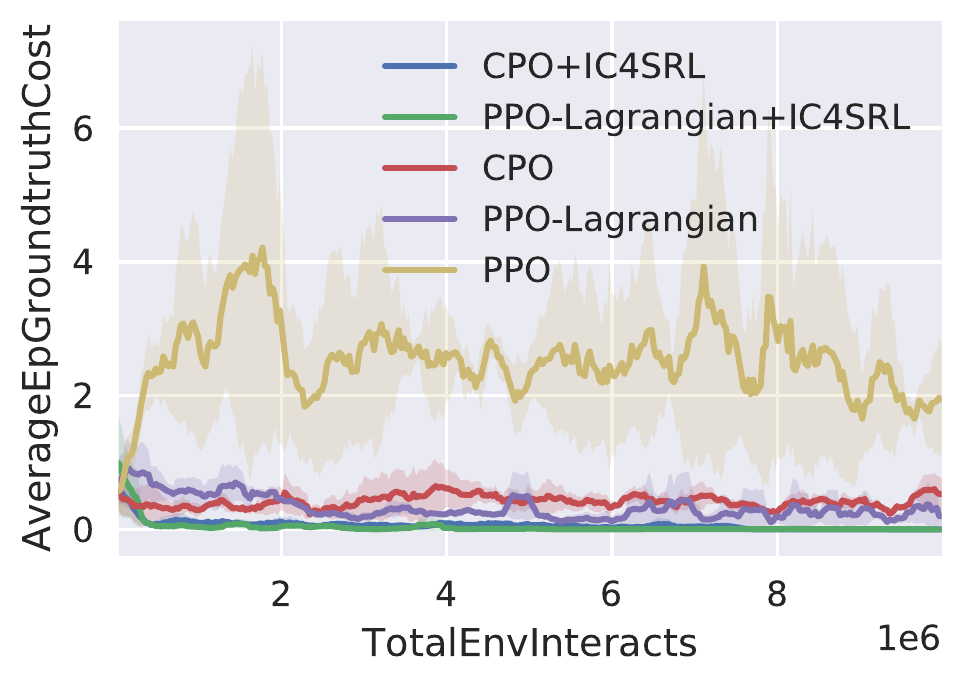}}
        \label{fig:push-pillar-point-Cost}
    \end{subfigure}
    \caption{Push-Pillar-Point}
    \label{Push-Pillar-Point}
    \end{subfigure}
    \hfill
    \vspace{-5pt}
    \caption{Average episodic return and extrinsic cost of IC4SRL and baseline methods on Safety Gym over five seeds. Goal-Hazard-Point is the training environment of AutoCost while the other three environments are unseen environments for IC4SRL.} 
    \label{fig:exp_generalize_different_tasks_and_constraints}
    \vspace{-15pt}
\end{figure*}
\subsection{Generalization Experiments}
\begin{figure}[htbp]
    \vspace{-5pt}
    \centering
    \begin{subfigure}[t]{0.23\textwidth}
    \begin{subfigure}[t]{1.00\textwidth}
        \raisebox{-\height}{\includegraphics[width=\textwidth]{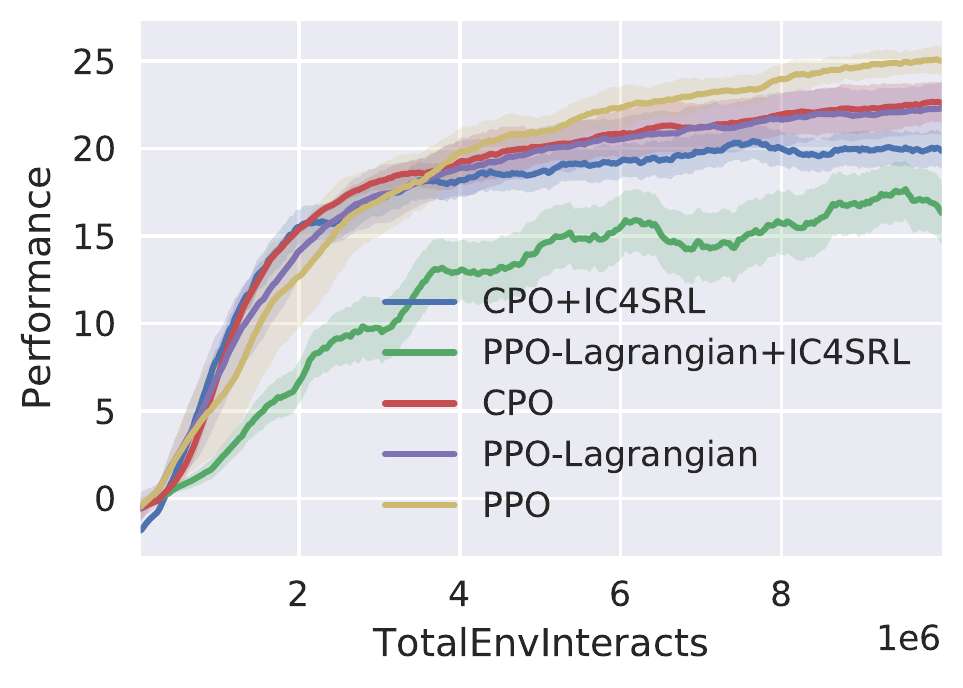}}
        \label{fig:goal-hazard-car-Performance}
    \end{subfigure}
    \hfill
    \begin{subfigure}[t]{1.00\textwidth}
        \raisebox{-\height}{\includegraphics[width=\textwidth]{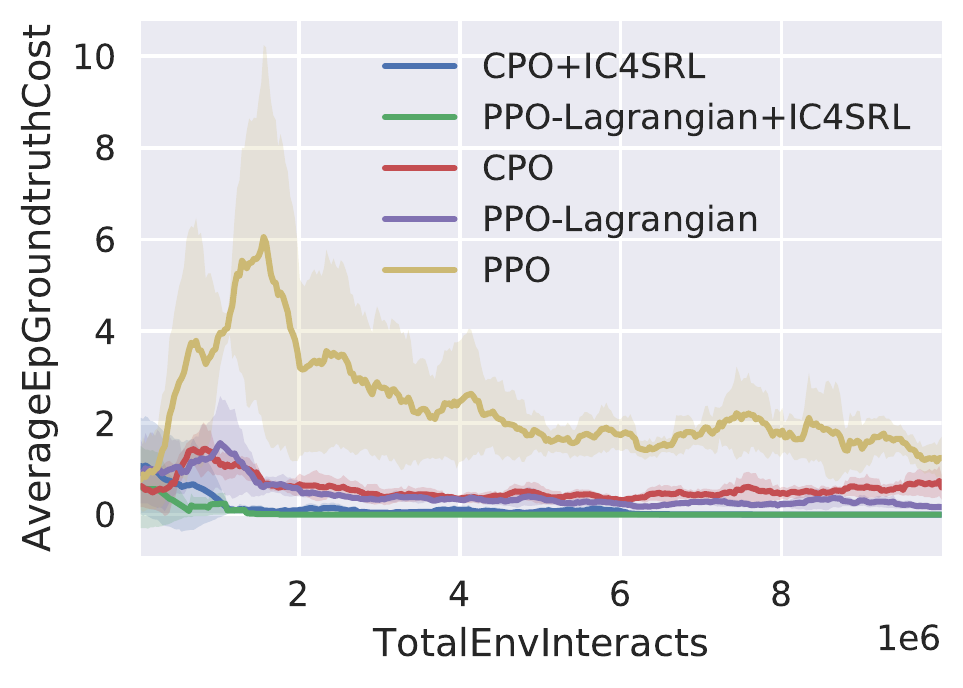}}
        \label{fig:goal-hazard-car-Cost}
    \end{subfigure}
    \caption{Goal-Hazard-Car}
    \label{Goal-Hazard-Car}
    \end{subfigure}
    \begin{subfigure}[t]{0.23\textwidth}
    \begin{subfigure}[t]{1.00\textwidth}
        \raisebox{-\height}{\includegraphics[width=\textwidth]{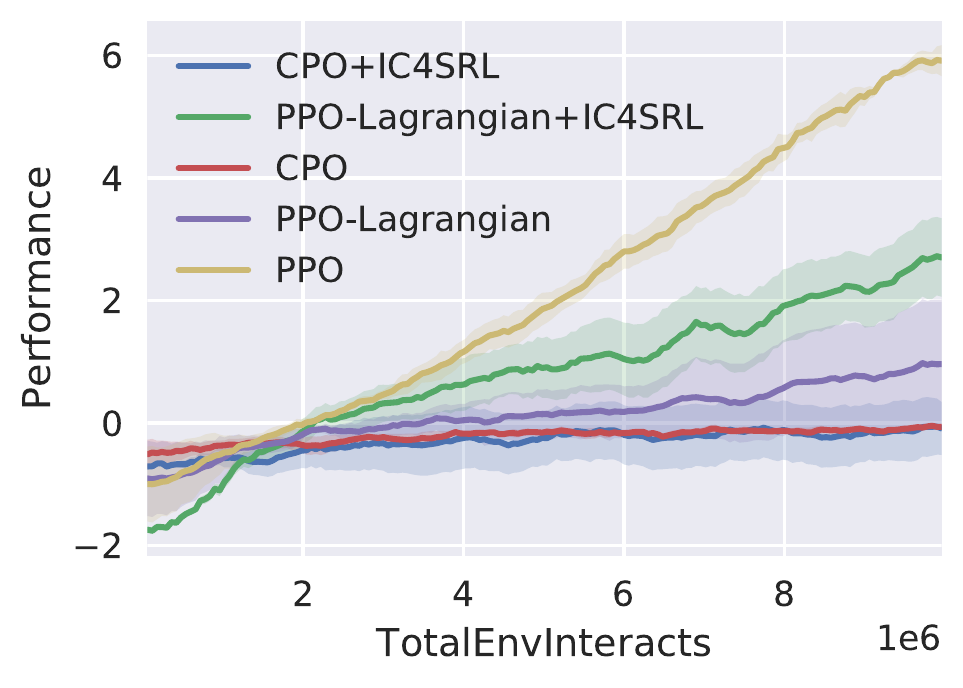}}
        \label{fig:goal-hazard-doggo-Performance}
    \end{subfigure}
    \hfill
    \begin{subfigure}[t]{1.00\textwidth}
        \raisebox{-\height}{\includegraphics[width=\textwidth]{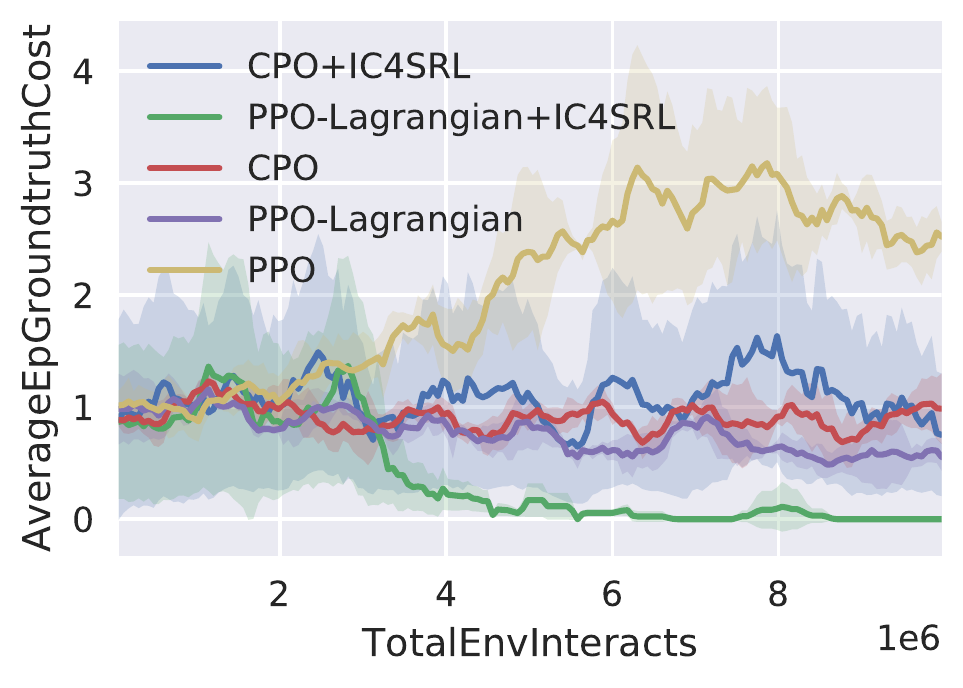}}
        \label{fig:goal-hazard-doggo-Cost}
    \end{subfigure}
    \caption{Goal-Hazard-Doggo}
    \label{Goal-Hazard-Doggo}
    \end{subfigure}
    \vspace{-5pt}
    \caption{Average episodic return, episodic extrinsic cost of constraints of IC4SRL and baseline methods on Safety Gym with more complex robots over five seeds.} 
    \label{fig:exp_generalize_different_robots}
    \vspace{-15pt}
\end{figure}

Among all the intrinsic cost functions achieving zero violation during the evolution, we finalize the candidate with the highest reward performance and name it IC4SRL (intrinsic cost for safe RL).
In this section, we further test the generalization ability of IC4SRL in different environments. The methods in the comparison group include: unconstrained RL algorithm \textbf{PPO}~\cite{schulman2017proximal} and constrained safe RL algorithms \textbf{PPO-Lagrangian}~\citep{chow2017lagrangian}, \textbf{CPO}~\cite{achiam2017cpo}. 
We denote \textbf{CPO+IC4SRL} and \textbf{PPO-Lagrangian+IC4SRL} as the corresponding safe RL methods with IC4SRL as intrinsic cost added to the original extrinsic cost defined in the CMDP. 
We set the cost limit to zero for all safe RL methods since we aim to avoid any constraint violation. For all experiments, we use the same neural network architectures and hyper-parameter (more details are provided in \Cref{appendix: experiment details}). 
\paragraph{Generalize to Different Tasks and Constraints} IC4SRL is searched from the \textit{training environment} \texttt{Goal-Hazard-Point}, and how well can the intrinsic cost function transfer to different \textit{test environments} remains a question.
To answer the question, we first test IC4SRL in different tasks and constraint types. The results are shown in \Cref{fig:exp_generalize_different_tasks_and_constraints}, where both CPO+IC4SRL and PPO-Lagrangian+IC4SRL quickly converge to zero-violation policies while CPO and PPO-Lagrangian without intrinsic cost fail to achieve zero violation. The difference between adding intrinsic cost and not adding intrinsic cost proves a proper cost function is a key factor towards achieving zero violation. We also notice that adding intrinsic cost also lowers the reward performance of both CPO and PPO-Lagrangian, indicating the trade-off between safety and performance where IC4SRL drives RL agents to learn more conservative behaviors to avoid any violation.
\paragraph{Generalize to Different Robots} To see whether IC4SRL is able to transfer to different robots, we replace the point robot with more complex robots like car and doggo. The evaluation results are shown in \Cref{fig:exp_generalize_different_robots}. Both CPO+IC4SRL and PPO-Lagrangian+IC4SRL achieve zero violation with the car robot, whereas only PPO-Lagrangian+IC4SRL converges to a zero-violation with the doggo robot. We find that both CPO and CPO+IC4SRL totally fail to gain meaningful rewards in the doggo environment, indicating the zero-violation policy brought by IC4SRL builds on the precondition that safe RL methods work (i.e., learn meaningful behaviors) in the environment.

\paragraph{The Limit of Generalization Ability} Though IC4SRL successfully generalizes to different environments as shown in \Cref{fig:exp_generalize_different_tasks_and_constraints} and \Cref{fig:exp_generalize_different_robots}, there is still a limit of its generalization ability. 
One limit is that IC4SRL is only searched with two safe RL algorithms. It is possible that some other safe RL methods require even more conservative intrinsic cost functions to achieve zero violation.
Another major limitation is that IC4SRL is coupled with the properties and features of Safety Gym. 
However, though IC4SRL may be not universal to all settings, one can always apply AutoCost to automatically search for a proper intrinsic cost function for either new safe RL algorithms or new environments.

\subsection{Visualization of Intrinsic Cost}
\begin{figure}[htbp]
    \vspace{-5pt}
    \begin{subfigure}[t]{0.23\textwidth}
        \centering
        {\includegraphics[width=\textwidth]{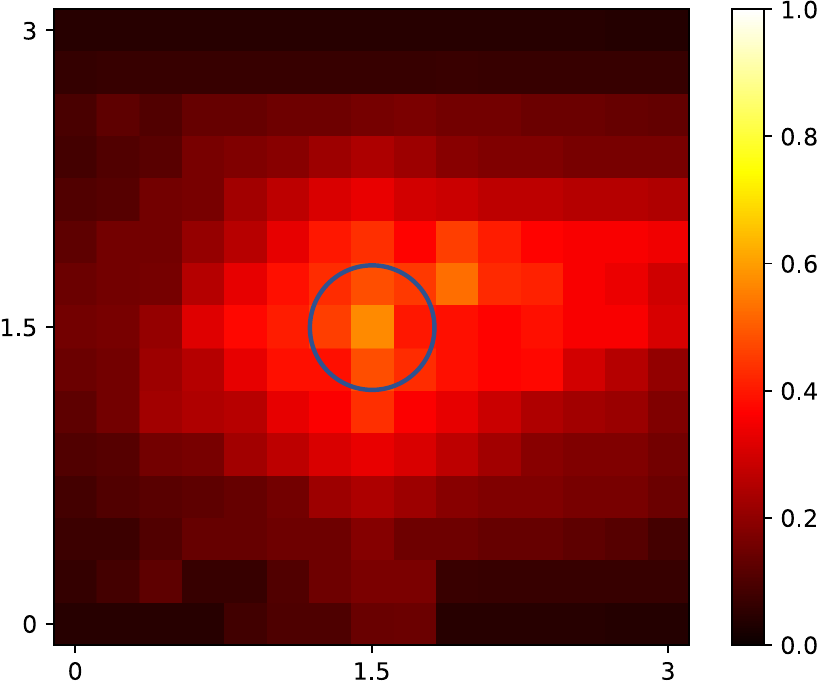}}
        \caption{Heatmap of IC4SRL}
        \label{fig:visualization_heatmap_intrinsic_cost}
    \end{subfigure}
    \begin{subfigure}[t]{0.23\textwidth}
        \centering
        {\includegraphics[width=\textwidth]{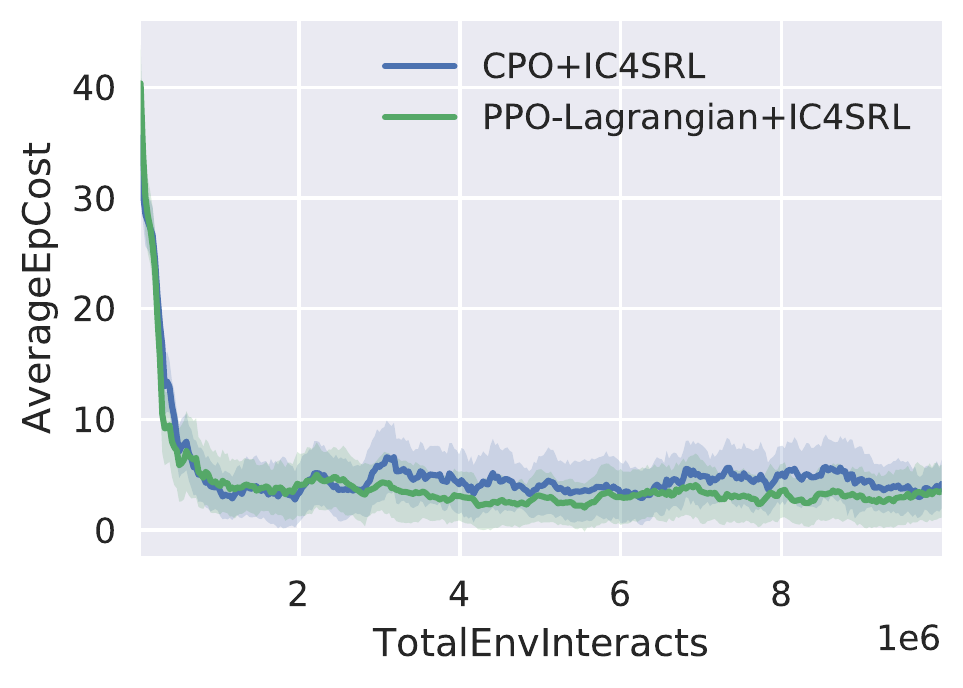}}
        \caption{$c^{ex+in}_\theta$}
        \label{fig:visualization_cost_curve}
    \end{subfigure}
    \vspace{-5pt}
    \caption{(a) Heatmap of intrinsic cost IC4SRL with different positions of the robot. The blue circle in the middle represents the hazard. (b) The average episodic sum of intrinsic cost and extrinsic cost for CPO+IC4SRL and PPO-Lagrangian+IC4SRL in Goal-Hazard-Point.}
    \label{fig:visualization}
    \vspace{-15pt}
\end{figure}

To further investigate what the searched intrinsic cost function capture, we visualize IC4SRL in \Cref{fig:visualization_heatmap_intrinsic_cost}. The blue circle in the middle represents the hazard. And each grid represents the positions of robots. The results show that IC4SRL assigns intrinsic cost values to the states next to the hazard, indicating that IC4SRL can capture local-aware information about the unsafe state surroundings. This result also echoes recent works~\citep{ma2021learn} which adds prior measures to the cost function to improve safety performance. Also the heatmap of intrinsic cost in \Cref{fig:visualization_heatmap_intrinsic_cost} shows a similar landscape of neural barrier functions~\citep{dawson2022safe}. But learning a neural barrier function requires pre-defined safe and unsafe regions in the state space, which is hard to design in complex environments. On the contrary, IC4SRL is discovered in a fully automated manner with no expert knowledge required.

To further illustrate why adding intrinsic cost is helpful for achieving zero-violation safe RL policies, we plot the training curve of $c^{ex+in}_\theta$ in \Cref{fig:visualization_cost_curve}. Note that both safe RL methods fail to lower $c^{ex+in}_\theta$ under the cost limit of zero. However, the ground truth extrinsic cost quickly converges to zero as shown in \Cref{Goal-Hazard-Point}. This indicates that adding a more conservative intrinsic cost like IC4SRL is the key to zero violation for safe RL algorithms. 
\subsection{The Necessity of AutoCost}
\begin{figure}[htbp]
    \vspace{-5pt}
    \begin{subfigure}[t]{0.23\textwidth}
        \centering
        {\includegraphics[width=\textwidth]{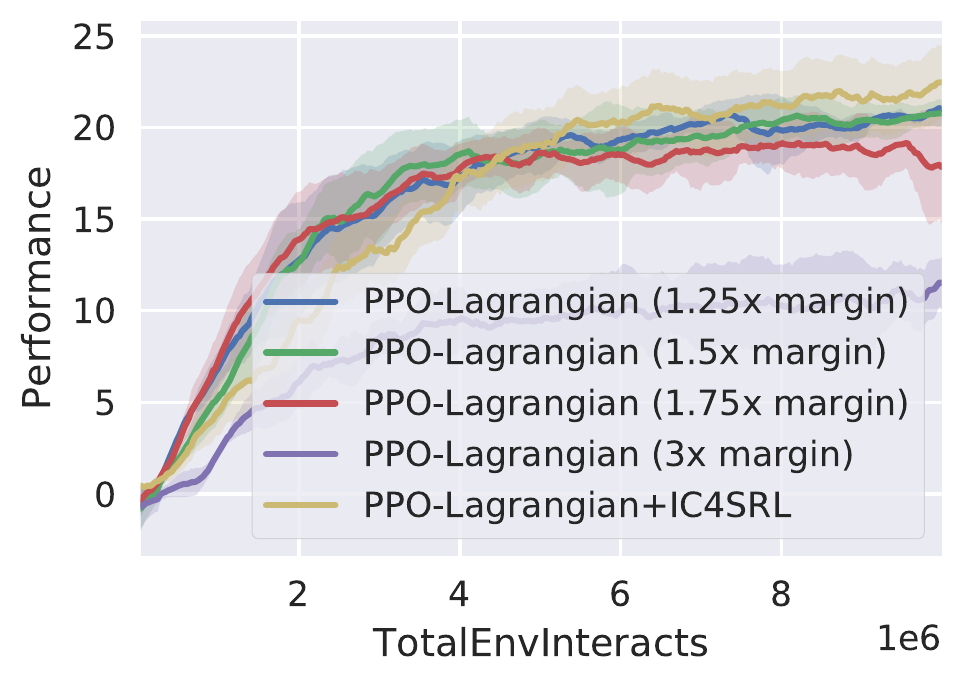}}
        \caption{Performance}
        \label{fig:safety_margin_performance_ppo-lag}
    \end{subfigure}
    \begin{subfigure}[t]{0.23\textwidth}
        \centering
        {\includegraphics[width=\textwidth]{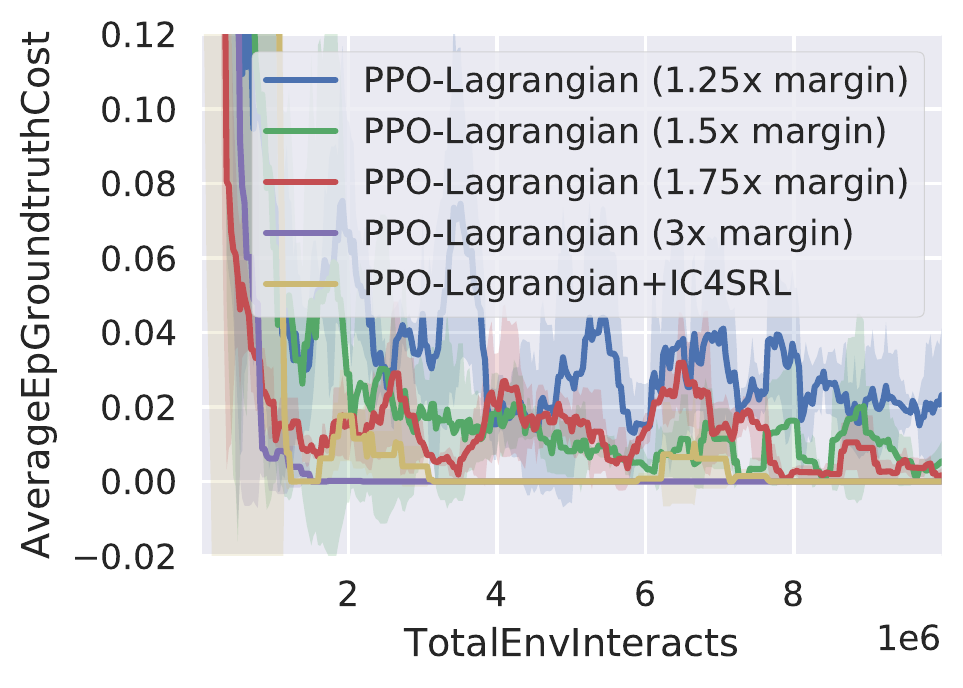}}
        \caption{Extrinsic cost}
        \label{fig:safety_margin_averageepgroundtruthcost_ppo-lag}
    \end{subfigure}
    \vspace{-5pt}
    \caption{Average episodic return, episodic extrinsic cost of PPO-Lagrangian with hand-tuned intrinsic cost functions with different safety margins.}
    \label{fig:safety_margin_ppo_lagrangian}
    \vspace{-15pt}
\end{figure}

The heatmap shown in \Cref{fig:visualization_heatmap_intrinsic_cost} demonstrates that intrinsic cost adds an extra margin to the constraint. To justify the necessity of AutoCost, we compare IC4SRL with hand-tuned intrinsic cost with different sizes of the safety margin. The results are shown in \Cref{fig:safety_margin_ppo_lagrangian}, where 3x margin and IC4SRL both obtain zero-violation policies after convergence. But IC4SRL achieves much better reward performance, which may be due to the \textit{selection} procedure where AutoCost selects zero-violation candidates with higher reward performance. Note that tuning a proper safety margin (e.g., 3x margin) requires many human efforts while IC4SRL is discovered in a fully automated manner. More importantly, hand-tuned intrinsic cost functions require much prior knowledge (e.g., the size of the constraint and the distances from the robot to the constraint at each time step), which is unrealistic in practice. Nevertheless, AutoCost discovered IC4SRL only with information from lidar sensors, indicating the generalizability of AutoCost to more complex CMDP and black-box safety constraints.

\subsection{With or Without Extrinsic Cost}
\begin{figure}[htbp]
    \vspace{-5pt}
    \begin{subfigure}[t]{0.23\textwidth}
        \centering
        {\includegraphics[width=\textwidth]{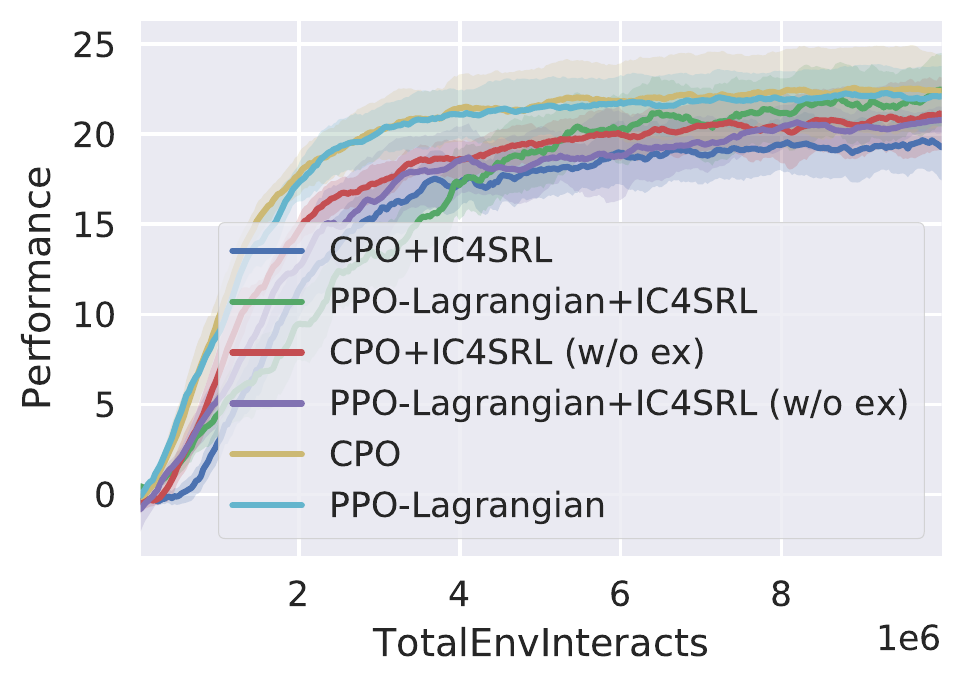}}
        \caption{Performance}
        \label{fig:with_or_without_extrinsic_reward}
    \end{subfigure}
    \begin{subfigure}[t]{0.23\textwidth}
        \centering
        {\includegraphics[width=\textwidth]{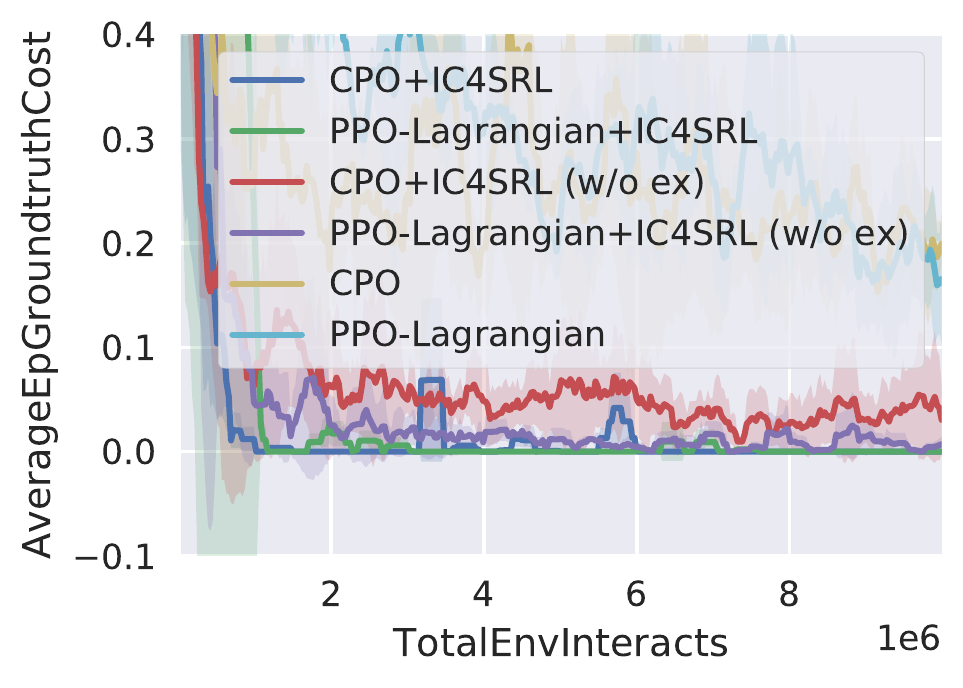}}
        \caption{Extrinsic cost}
        \label{fig:with_or_without_extrinsic_cost}
    \end{subfigure}
    \vspace{-5pt}
    \caption{Average episodic return, episodic extrinsic cost of CPO and PPO-Lagrangian with three different cost functions designs: (i) IC4SRL adds intrinsic cost to the extrinsic cost; (ii) IC4SRL (w/o ex) only uses intrinsic cost as the cost function; (iii) vanilla version only uses extrinsic cost.}
    \vspace{-15pt}
    \label{fig:with_or_without_extrinsic}
\end{figure}

To test whether safe RL methods are capable of achieving zero violation solely with the intrinsic cost, we denote the safe RL baselines without extrinsic cost function as \textbf{CPO+IC4SRL (w/o ex)} and \textbf{PPO-Lagrangian+IC4SRL (w/o ex)}. The results are shown in \Cref{fig:with_or_without_extrinsic}. Note that safe RL baselines without extrinsic cost eliminate much more violations compared with baselines using only the extrinsic cost function, which indicates the advantage of IC4SRL over naive indicator cost functions. But both CPO+IC4SRL (w/o ex) and PPO-Lagrangian+IC4SRL (w/o ex) fail to converge to a zero-violation policy. This may be due to the evolutionary process of AutoCost adopting the setting that the cost function is the sum of intrinsic and extrinsic costs and therefore the searched intrinsic cost is limited to that setting.

\section{Conclusion}
In this paper, we present AutoCost, a principled and universal framework for automated intrinsic cost design for safe RL. This is the first such framework to the best of our knowledge. By searching on Safety Gym with this framework, we discover IC4SRL, a top-performing intrinsic cost function that generalizes well to diverse test environments. Our empirical results show promise in using intrinsic cost function to achieve zero-violation safety performance with safe RL methods like CPO and Lagrangian methods. 
We hope our studies provide insights that will deepen the understanding of cost function design in safe RL, and shed light on how to enable avenues for real-world deployment of RL in areas like robotics where zero-violation safety is critical.


\bibliography{ref}  

\begin{thebibliography}{51}
\providecommand{\natexlab}[1]{#1}

\bibitem[{Achiam et~al.(2017)Achiam, Held, Tamar, and Abbeel}]{achiam2017cpo}
Achiam, J.; Held, D.; Tamar, A.; and Abbeel, P. 2017.
\newblock Constrained policy optimization.
\newblock In \emph{International conference on machine learning}, 22--31. PMLR.

\bibitem[{Ahmadi and Majumdar(2016)}]{ahmadi2016some}
Ahmadi, A.~A.; and Majumdar, A. 2016.
\newblock Some applications of polynomial optimization in operations research
  and real-time decision making.
\newblock \emph{Optimization Letters}, 10(4): 709--729.

\bibitem[{Altman(1999)}]{altman1999cmdp}
Altman, E. 1999.
\newblock \emph{Constrained Markov decision processes: stochastic modeling}.
\newblock Routledge.

\bibitem[{Ames, Grizzle, and Tabuada(2014)}]{ames2014control}
Ames, A.~D.; Grizzle, J.~W.; and Tabuada, P. 2014.
\newblock Control barrier function based quadratic programs with application to
  adaptive cruise control.
\newblock In \emph{53rd IEEE Conference on Decision and Control}, 6271--6278.
  IEEE.

\bibitem[{Andrychowicz et~al.(2017)Andrychowicz, Wolski, Ray, Schneider, Fong,
  Welinder, McGrew, Tobin, Pieter~Abbeel, and
  Zaremba}]{andrychowicz2017hindsight}
Andrychowicz, M.; Wolski, F.; Ray, A.; Schneider, J.; Fong, R.; Welinder, P.;
  McGrew, B.; Tobin, J.; Pieter~Abbeel, O.; and Zaremba, W. 2017.
\newblock Hindsight experience replay.
\newblock \emph{Advances in neural information processing systems}, 30.

\bibitem[{B{\"a}ck and Schwefel(1993)}]{back1993overview}
B{\"a}ck, T.; and Schwefel, H.-P. 1993.
\newblock An overview of evolutionary algorithms for parameter optimization.
\newblock \emph{Evolutionary computation}, 1(1): 1--23.

\bibitem[{Berkenkamp et~al.(2017)Berkenkamp, Turchetta, Schoellig, and
  Krause}]{berkenkamp2017safe}
Berkenkamp, F.; Turchetta, M.; Schoellig, A.; and Krause, A. 2017.
\newblock Safe model-based reinforcement learning with stability guarantees.
\newblock \emph{Advances in neural information processing systems}, 30.

\bibitem[{Bharadhwaj et~al.(2020)Bharadhwaj, Kumar, Rhinehart, Levine, Shkurti,
  and Garg}]{bharadhwaj2020conservative}
Bharadhwaj, H.; Kumar, A.; Rhinehart, N.; Levine, S.; Shkurti, F.; and Garg, A.
  2020.
\newblock Conservative safety critics for exploration.
\newblock \emph{arXiv preprint arXiv:2010.14497}.

\bibitem[{Boyd, Boyd, and Vandenberghe(2004)}]{boyd2004convex}
Boyd, S.; Boyd, S.~P.; and Vandenberghe, L. 2004.
\newblock \emph{Convex optimization}.
\newblock Cambridge university press.

\bibitem[{Brown and Sandholm(2018)}]{brown2018superhuman}
Brown, N.; and Sandholm, T. 2018.
\newblock Superhuman AI for heads-up no-limit poker: Libratus beats top
  professionals.
\newblock \emph{Science}, 359(6374): 418--424.

\bibitem[{Chang, Roohi, and Gao(2019)}]{chang2019neural}
Chang, Y.-C.; Roohi, N.; and Gao, S. 2019.
\newblock Neural lyapunov control.
\newblock \emph{Advances in neural information processing systems}, 32.

\bibitem[{Chen, Dong, and Wang(2021)}]{chen2021primal}
Chen, Y.; Dong, J.; and Wang, Z. 2021.
\newblock A primal-dual approach to constrained markov decision processes.
\newblock \emph{arXiv preprint arXiv:2101.10895}.

\bibitem[{Chow et~al.(2017)Chow, Ghavamzadeh, Janson, and
  Pavone}]{chow2017lagrangian}
Chow, Y.; Ghavamzadeh, M.; Janson, L.; and Pavone, M. 2017.
\newblock Risk-constrained reinforcement learning with percentile risk
  criteria.
\newblock \emph{The Journal of Machine Learning Research}, 18(1): 6070--6120.

\bibitem[{Chow et~al.(2018)Chow, Nachum, Duenez-Guzman, and
  Ghavamzadeh}]{chow2018lyapunov}
Chow, Y.; Nachum, O.; Duenez-Guzman, E.; and Ghavamzadeh, M. 2018.
\newblock A lyapunov-based approach to safe reinforcement learning.
\newblock \emph{Advances in neural information processing systems}, 31.

\bibitem[{Co-Reyes et~al.(2020)Co-Reyes, Miao, Peng, Real, Le, Levine, Lee, and
  Faust}]{co2020evolvingRL}
Co-Reyes, J.~D.; Miao, Y.; Peng, D.; Real, E.; Le, Q.~V.; Levine, S.; Lee, H.;
  and Faust, A. 2020.
\newblock Evolving Reinforcement Learning Algorithms.
\newblock In \emph{International Conference on Learning Representations}.

\bibitem[{Dalal et~al.(2018)Dalal, Dvijotham, Vecerik, Hester, Paduraru, and
  Tassa}]{dalal2018safe}
Dalal, G.; Dvijotham, K.; Vecerik, M.; Hester, T.; Paduraru, C.; and Tassa, Y.
  2018.
\newblock Safe exploration in continuous action spaces.
\newblock \emph{arXiv preprint arXiv:1801.08757}.

\bibitem[{Dawson, Gao, and Fan(2022)}]{dawson2022survey}
Dawson, C.; Gao, S.; and Fan, C. 2022.
\newblock Safe Control with Learned Certificates: A Survey of Neural Lyapunov,
  Barrier, and Contraction methods.
\newblock \emph{arXiv preprint arXiv:2202.11762}.

\bibitem[{Dawson et~al.(2022)Dawson, Qin, Gao, and Fan}]{dawson2022safe}
Dawson, C.; Qin, Z.; Gao, S.; and Fan, C. 2022.
\newblock Safe nonlinear control using robust neural lyapunov-barrier
  functions.
\newblock In \emph{Conference on Robot Learning}, 1724--1735. PMLR.

\bibitem[{Dewey(2014)}]{dewey2014reinforcement}
Dewey, D. 2014.
\newblock Reinforcement learning and the reward engineering principle.
\newblock In \emph{2014 AAAI Spring Symposium Series}.

\bibitem[{El~Chamie, Yu, and
  A{\c{c}}{\i}kme{\c{s}}e(2016)}]{el2016convexsynthesis}
El~Chamie, M.; Yu, Y.; and A{\c{c}}{\i}kme{\c{s}}e, B. 2016.
\newblock Convex synthesis of randomized policies for controlled Markov chains
  with density safety upper bound constraints.
\newblock In \emph{2016 American Control Conference (ACC)}, 6290--6295. IEEE.

\bibitem[{Faust, Francis, and Mehta(2019)}]{faust2019evolvingreward}
Faust, A.; Francis, A.; and Mehta, D. 2019.
\newblock Evolving rewards to automate reinforcement learning.
\newblock \emph{arXiv preprint arXiv:1905.07628}.

\bibitem[{Ferlez et~al.(2020)Ferlez, Elnaggar, Shoukry, and
  Fleming}]{ferlez2020shieldnn}
Ferlez, J.; Elnaggar, M.; Shoukry, Y.; and Fleming, C. 2020.
\newblock Shieldnn: A provably safe nn filter for unsafe nn controllers.
\newblock \emph{CoRR}, abs/2006.09564.

\bibitem[{Franke et~al.(2020)Franke, K{\"o}hler, Biedenkapp, and
  Hutter}]{franke2020sample}
Franke, J.~K.; K{\"o}hler, G.; Biedenkapp, A.; and Hutter, F. 2020.
\newblock Sample-efficient automated deep reinforcement learning.
\newblock \emph{arXiv preprint arXiv:2009.01555}.

\bibitem[{Giesl and Hafstein(2015)}]{giesl2015review}
Giesl, P.; and Hafstein, S. 2015.
\newblock Review on computational methods for Lyapunov functions.
\newblock \emph{Discrete \& Continuous Dynamical Systems-B}, 20(8): 2291.

\bibitem[{Gracia, Garelli, and Sala(2013)}]{gracia2013reactive}
Gracia, L.; Garelli, F.; and Sala, A. 2013.
\newblock Reactive sliding-mode algorithm for collision avoidance in robotic
  systems.
\newblock \emph{IEEE Transactions on Control Systems Technology}, 21(6):
  2391--2399.

\bibitem[{Hong et~al.(2018)Hong, Shann, Su, Chang, Fu, and
  Lee}]{hong2018diversity}
Hong, Z.-W.; Shann, T.-Y.; Su, S.-Y.; Chang, Y.-H.; Fu, T.-J.; and Lee, C.-Y.
  2018.
\newblock Diversity-driven exploration strategy for deep reinforcement
  learning.
\newblock \emph{Advances in neural information processing systems}, 31.

\bibitem[{Houthooft et~al.(2018)Houthooft, Chen, Isola, Stadie, Wolski,
  Jonathan~Ho, and Abbeel}]{houthooft2018evolvedPG}
Houthooft, R.; Chen, Y.; Isola, P.; Stadie, B.; Wolski, F.; Jonathan~Ho, O.;
  and Abbeel, P. 2018.
\newblock Evolved policy gradients.
\newblock \emph{Advances in Neural Information Processing Systems}, 31.

\bibitem[{Khatib(1986)}]{khatib1986real}
Khatib, O. 1986.
\newblock Real-time obstacle avoidance for manipulators and mobile robots.
\newblock In \emph{Autonomous robot vehicles}, 396--404. Springer.

\bibitem[{Kong et~al.(2015)Kong, Pfeiffer, Schildbach, and
  Borrelli}]{kong2015kinematic}
Kong, J.; Pfeiffer, M.; Schildbach, G.; and Borrelli, F. 2015.
\newblock Kinematic and dynamic vehicle models for autonomous driving control
  design.
\newblock In \emph{2015 IEEE Intelligent Vehicles Symposium (IV)}, 1094--1099.
  IEEE.

\bibitem[{Liang, Que, and Modiano(2018)}]{liang2018accelerated}
Liang, Q.; Que, F.; and Modiano, E. 2018.
\newblock Accelerated primal-dual policy optimization for safe reinforcement
  learning.
\newblock \emph{arXiv preprint arXiv:1802.06480}.

\bibitem[{Liu and Tomizuka(2014)}]{liu2014control}
Liu, C.; and Tomizuka, M. 2014.
\newblock Control in a safe set: Addressing safety in human-robot interactions.
\newblock In \emph{Dynamic Systems and Control Conference}, volume 46209,
  V003T42A003. American Society of Mechanical Engineers.

\bibitem[{Ma et~al.(2021)Ma, Liu, Li, Zheng, Sun, and Chen}]{ma2021learn}
Ma, H.; Liu, C.; Li, S.~E.; Zheng, S.; Sun, W.; and Chen, J. 2021.
\newblock Learn Zero-Constraint-Violation Policy in Model-Free Constrained
  Reinforcement Learning.
\newblock \emph{arXiv preprint arXiv:2111.12953}.

\bibitem[{Mnih et~al.(2015)Mnih, Kavukcuoglu, Silver, Rusu, Veness, Bellemare,
  Graves, Riedmiller, Fidjeland, Ostrovski et~al.}]{mnih2015human}
Mnih, V.; Kavukcuoglu, K.; Silver, D.; Rusu, A.~A.; Veness, J.; Bellemare,
  M.~G.; Graves, A.; Riedmiller, M.; Fidjeland, A.~K.; Ostrovski, G.; et~al.
  2015.
\newblock Human-level control through deep reinforcement learning.
\newblock \emph{nature}, 518(7540): 529--533.

\bibitem[{Paul, Kurin, and Whiteson(2019)}]{paul2019fast}
Paul, S.; Kurin, V.; and Whiteson, S. 2019.
\newblock Fast efficient hyperparameter tuning for policy gradient methods.
\newblock \emph{Advances in Neural Information Processing Systems}, 32.

\bibitem[{Qin et~al.(2021)Qin, Zhang, Chen, Chen, and Fan}]{qin2021learning}
Qin, Z.; Zhang, K.; Chen, Y.; Chen, J.; and Fan, C. 2021.
\newblock Learning safe multi-agent control with decentralized neural barrier
  certificates.
\newblock \emph{arXiv preprint arXiv:2101.05436}.

\bibitem[{Ray, Achiam, and Amodei(2019)}]{ray2019safetygym}
Ray, A.; Achiam, J.; and Amodei, D. 2019.
\newblock Benchmarking safe exploration in deep reinforcement learning.
\newblock \emph{CoRR}, abs/1910.01708.

\bibitem[{Runge et~al.(2018)Runge, Stoll, Falkner, and Hutter}]{runge2018rna}
Runge, F.; Stoll, D.; Falkner, S.; and Hutter, F. 2018.
\newblock Learning to design RNA.
\newblock \emph{arXiv preprint arXiv:1812.11951}.

\bibitem[{Schulman et~al.(2015)Schulman, Levine, Abbeel, Jordan, and
  Moritz}]{schulman2015trpo}
Schulman, J.; Levine, S.; Abbeel, P.; Jordan, M.; and Moritz, P. 2015.
\newblock Trust region policy optimization.
\newblock In \emph{International conference on machine learning}, 1889--1897.
  PMLR.

\bibitem[{Schulman et~al.(2017)Schulman, Wolski, Dhariwal, Radford, and
  Klimov}]{schulman2017proximal}
Schulman, J.; Wolski, F.; Dhariwal, P.; Radford, A.; and Klimov, O. 2017.
\newblock Proximal Policy Optimization Algorithms.
\newblock \emph{CoRR}, abs/1707.06347.

\bibitem[{Srinivasan et~al.(2020)Srinivasan, Eysenbach, Ha, Tan, and
  Finn}]{srinivasan2020learningtobesafe}
Srinivasan, K.; Eysenbach, B.; Ha, S.; Tan, J.; and Finn, C. 2020.
\newblock Learning to be safe: Deep rl with a safety critic.
\newblock \emph{arXiv preprint arXiv:2010.14603}.

\bibitem[{Stooke, Achiam, and Abbeel(2020)}]{stooke2020responsive}
Stooke, A.; Achiam, J.; and Abbeel, P. 2020.
\newblock Responsive safety in reinforcement learning by pid lagrangian
  methods.
\newblock In \emph{International Conference on Machine Learning}, 9133--9143.
  PMLR.

\bibitem[{Tessler, Mankowitz, and Mannor(2018)}]{tessler2018reward}
Tessler, C.; Mankowitz, D.~J.; and Mannor, S. 2018.
\newblock Reward constrained policy optimization.
\newblock \emph{arXiv preprint arXiv:1805.11074}.

\bibitem[{Todorov, Erez, and Tassa(2012)}]{todorov2012mujoco}
Todorov, E.; Erez, T.; and Tassa, Y. 2012.
\newblock Mujoco: A physics engine for model-based control.
\newblock In \emph{2012 IEEE/RSJ international conference on intelligent robots
  and systems}, 5026--5033. IEEE.

\bibitem[{Veeriah et~al.(2019)Veeriah, Hessel, Xu, Rajendran, Lewis, Oh, van
  Hasselt, Silver, and Singh}]{veeriah2019discovery}
Veeriah, V.; Hessel, M.; Xu, Z.; Rajendran, J.; Lewis, R.~L.; Oh, J.; van
  Hasselt, H.~P.; Silver, D.; and Singh, S. 2019.
\newblock Discovery of useful questions as auxiliary tasks.
\newblock \emph{Advances in Neural Information Processing Systems}, 32.

\bibitem[{Vinyals et~al.(2019)Vinyals, Babuschkin, Czarnecki, Mathieu, Dudzik,
  Chung, Choi, Powell, Ewalds, Georgiev et~al.}]{vinyals2019grandmaster}
Vinyals, O.; Babuschkin, I.; Czarnecki, W.~M.; Mathieu, M.; Dudzik, A.; Chung,
  J.; Choi, D.~H.; Powell, R.; Ewalds, T.; Georgiev, P.; et~al. 2019.
\newblock Grandmaster level in StarCraft II using multi-agent reinforcement
  learning.
\newblock \emph{Nature}, 575(7782): 350--354.

\bibitem[{Wei and Liu(2019)}]{wei2019safe}
Wei, T.; and Liu, C. 2019.
\newblock Safe control algorithms using energy functions: A uni ed framework,
  benchmark, and new directions.
\newblock In \emph{2019 IEEE 58th Conference on Decision and Control (CDC)},
  238--243. IEEE.

\bibitem[{Xu et~al.(2020)Xu, van Hasselt, Hessel, Oh, Singh, and
  Silver}]{xu2020meta}
Xu, Z.; van Hasselt, H.~P.; Hessel, M.; Oh, J.; Singh, S.; and Silver, D. 2020.
\newblock Meta-gradient reinforcement learning with an objective discovered
  online.
\newblock \emph{Advances in Neural Information Processing Systems}, 33:
  15254--15264.

\bibitem[{Zahavy et~al.(2020)Zahavy, Xu, Veeriah, Hessel, Oh, van Hasselt,
  Silver, and Singh}]{zahavy2020self}
Zahavy, T.; Xu, Z.; Veeriah, V.; Hessel, M.; Oh, J.; van Hasselt, H.~P.;
  Silver, D.; and Singh, S. 2020.
\newblock A self-tuning actor-critic algorithm.
\newblock \emph{Advances in Neural Information Processing Systems}, 33:
  20913--20924.

\bibitem[{Zhang, Vuong, and Ross(2020)}]{zhang2020first}
Zhang, Y.; Vuong, Q.; and Ross, K. 2020.
\newblock First order constrained optimization in policy space.
\newblock \emph{Advances in Neural Information Processing Systems}, 33:
  15338--15349.

\bibitem[{Zhao, He, and Liu(2021)}]{zhao2021issa}
Zhao, W.; He, T.; and Liu, C. 2021.
\newblock Model-free safe control for zero-violation reinforcement learning.
\newblock In \emph{5th Annual Conference on Robot Learning}.

\bibitem[{Zheng, Oh, and Singh(2018)}]{zheng2018intrinsic}
Zheng, Z.; Oh, J.; and Singh, S. 2018.
\newblock On learning intrinsic rewards for policy gradient methods.
\newblock \emph{Advances in Neural Information Processing Systems}, 31.

\end{thebibliography}

\clearpage
\appendix
\section{More Experiment Results}
\subsection{More Hand-deigned Cost Functions}
\label{appendix:manually_designed}
\begin{figure}[H]
    \begin{subfigure}[t]{0.23\textwidth}
        \centering
        {\includegraphics[width=\textwidth]{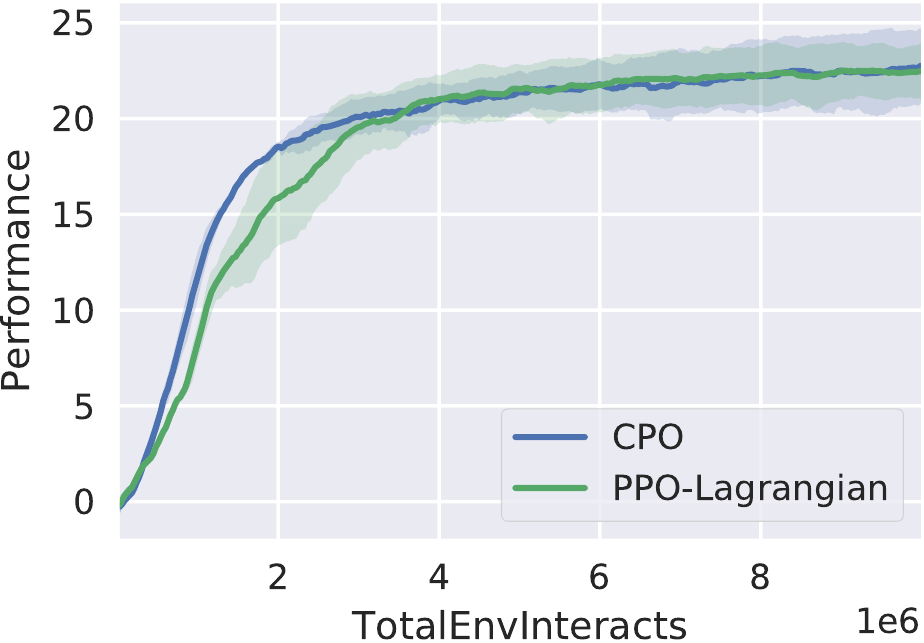}}
        \caption{Performance}
        \label{fig:appendix_delta_d_change_performance}
    \end{subfigure}
    \begin{subfigure}[t]{0.23\textwidth}
        \centering
        {\includegraphics[width=\textwidth]{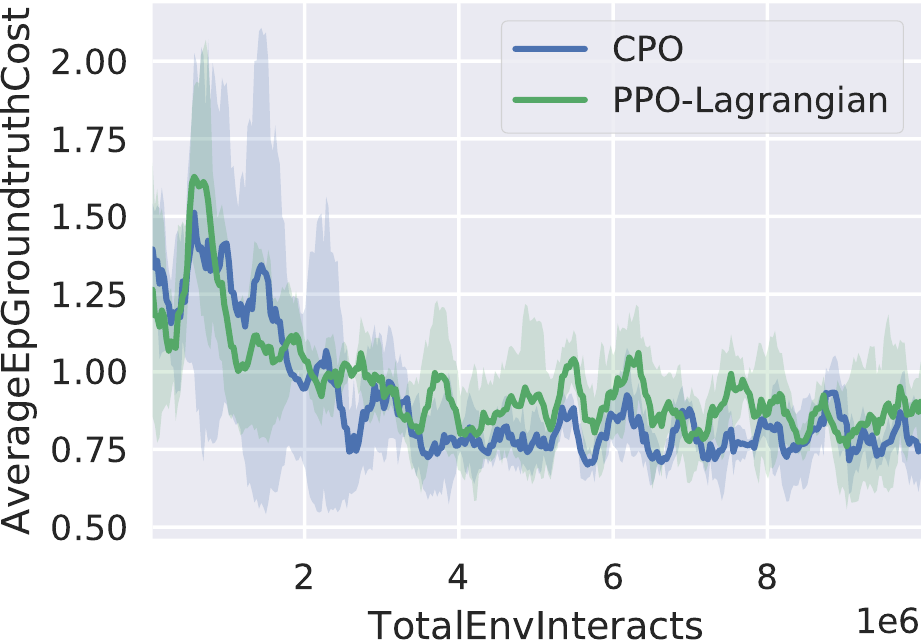}}
        \caption{Extrinsic cost}
        \label{fig:appendix_delta_d_change_cost}
    \end{subfigure}
    \vspace{-5pt}
    \caption{Average episodic reward and cost of safe RL baselines with a cost function based on the change of distance.}
    \label{fig:appendix_delta_d_change}
    \vspace{-15pt}
\end{figure}

In this subsection, we present more experiment results of several other hand-designed cost functions to illustrate the difficulty of manually designing cost functions.
\paragraph{Distance Change}
We design a cost function based on the change of distance between two time steps:
\begin{equation}
c(s_t) = d(s_{t-1}) - d(s_t)~,
\label{eq:appendix_delta_d_change}
\end{equation}
where $d(\cdot)$ denotes the distance from RL agent to the closet constraint. Note that the difference between \Cref{eq:appendix_delta_d_change} and \Cref{eq:empirical_dense_cost} is whether to enforce the cost function to be non-negative. We train CPO and PPO-Lagrangian under the denser cost function. The results are shown in \Cref{fig:appendix_delta_d_change}, where both CPO and PPO-Lagrangian fail to lower the average episodic extrinsic cost to zero. The main reason for such failure is that the cost function defined in \Cref{eq:appendix_delta_d_change} could be negative, which make the zero cost limit invalid for most scenarios. This result proves the necessity of a non-negative cost function.

\begin{figure}[H]
    \begin{subfigure}[t]{0.23\textwidth}
        \centering
        {\includegraphics[width=\textwidth]{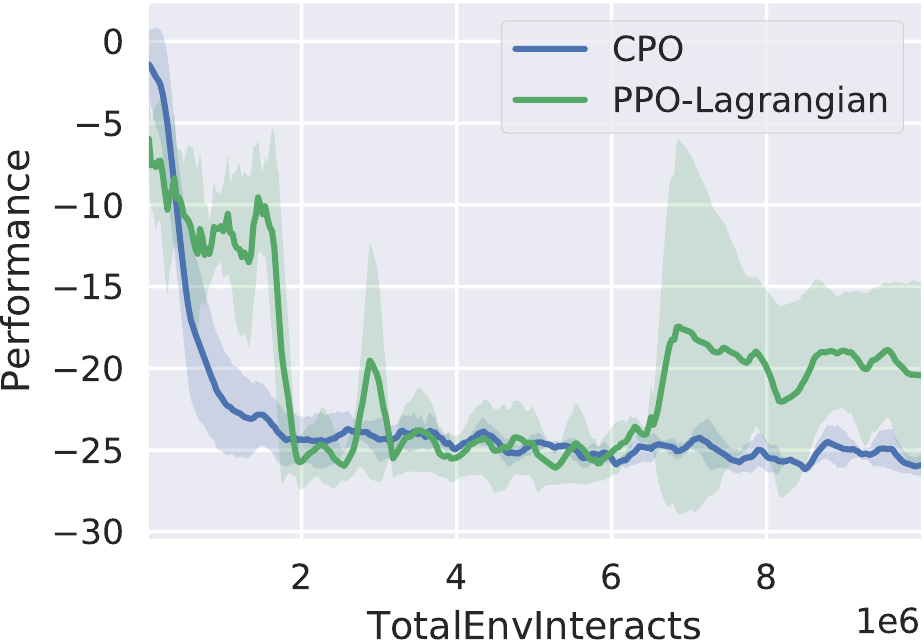}}
        \caption{Performance}
        \label{fig:appendix_delta_d_binary_performance}
    \end{subfigure}
    \begin{subfigure}[t]{0.23\textwidth}
        \centering
        {\includegraphics[width=\textwidth]{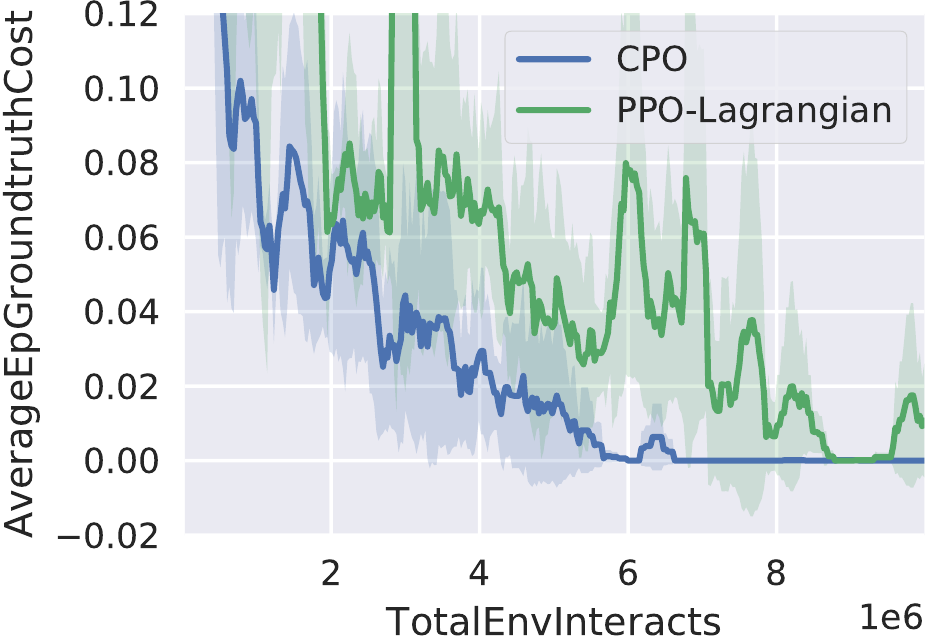}}
        \caption{Extrinsic cost}
        \label{fig:appendix_delta_d_binary_cost}
    \end{subfigure}
    \vspace{-5pt}
    \caption{Average episodic reward and cost of safe RL baselines with a cost function based on the change of distance.}
    \label{fig:appendix_delta_d_binary}
    \vspace{-15pt}
\end{figure}

\paragraph{Indicator Distance Change}
We design a cost function based on the change of distance between two time steps:
\begin{equation}
c(s_t) = \mathbbm{1}\Big[d(s_{t-1}) - d(s_t) \geq 0 \Big]~,
\label{eq:appendix_delta_d_binary}
\end{equation}
where $d(\cdot)$ denotes the distance from RL agent to the closet constraint. Note the cost function assigns $1$ to any step getting closer to the constraint and $0$ otherwise. We train CPO and PPO-Lagrangian under the cost function defined in \Cref{eq:appendix_delta_d_binary}. The results are shown in \Cref{fig:appendix_delta_d_binary}, where both CPO converges to a zero-violation policy while PPO-Lagrangian obtains a near zero-violation policy after convergence. But both CPO and PPO-Lagrangian suffer from terrible reward performance drop. 

\subsection{CPO with Different Safety Margins}
\begin{figure}[htbp]
    \begin{subfigure}[t]{0.23\textwidth}
        \centering
        {\includegraphics[width=\textwidth]{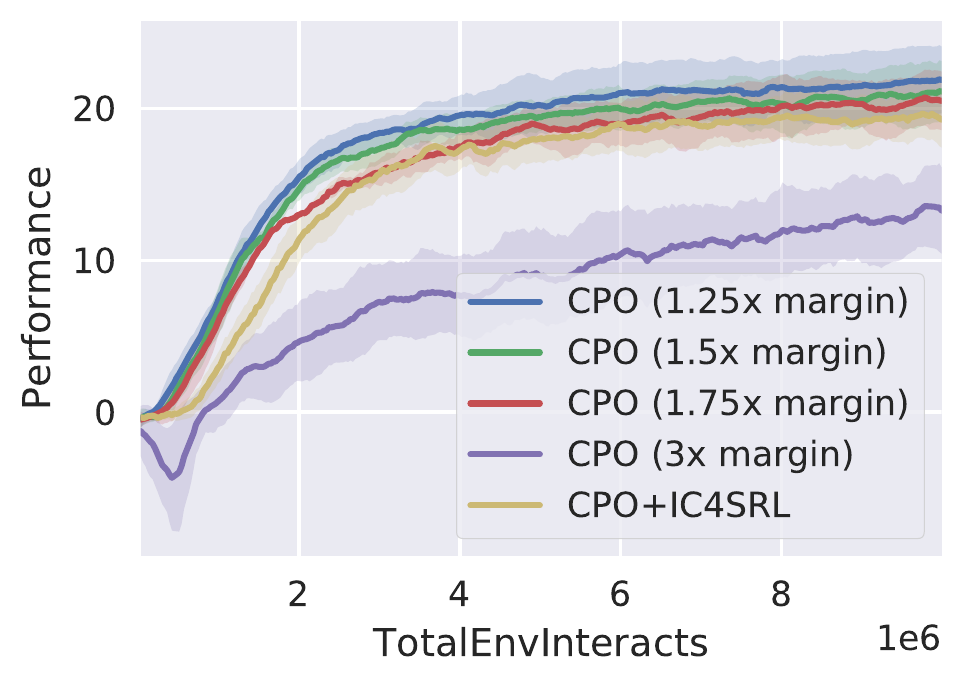}}
        \caption{Performance}
        \label{fig:safety_margin_performance_cpo}
    \end{subfigure}
    \begin{subfigure}[t]{0.23\textwidth}
        \centering
        {\includegraphics[width=\textwidth]{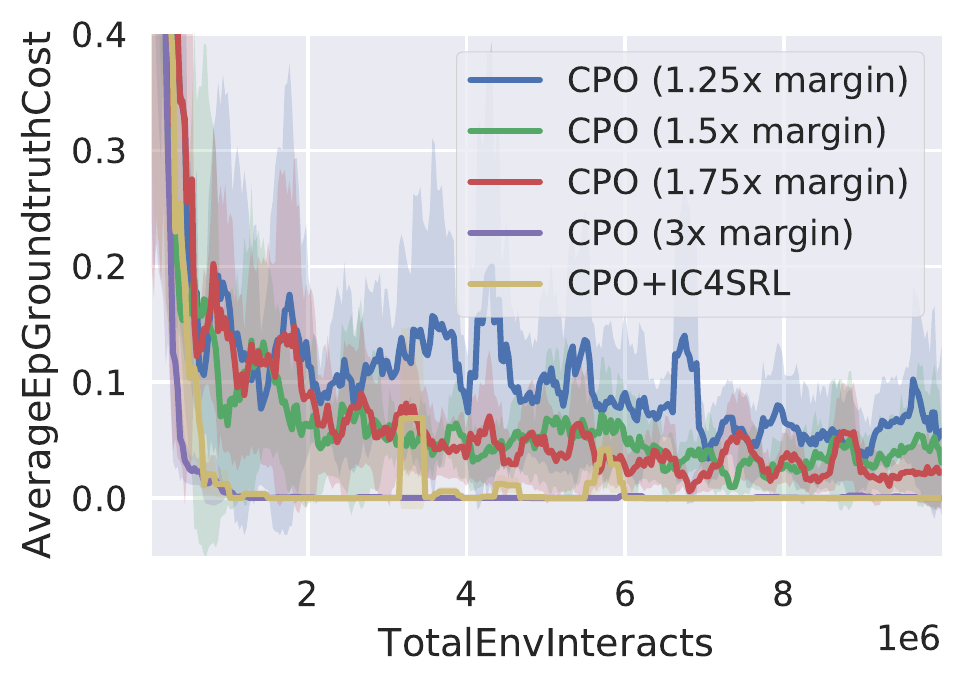}}
        \caption{Extrinsic cost}
        \label{fig:safety_margin_averageepgroundtruthcost_cpo}
    \end{subfigure}
    \vspace{-5pt}
    \caption{Average episodic return, episodic extrinsic cost of PPO-Lagrangian with hand-tuned intrinsic cost functions with different safety margins.}
    \label{fig:safety_margin_cpo}
    \vspace{-15pt}
\end{figure}

To justify the necessity of AutoCost, we compare IC4SRL with hand-tuned intrinsic cost with different sizes of the safety margin. As shown in \Cref{fig:safety_margin_cpo}, the results of CPO are similar to PPO-Lagrangian, where 3x margin and IC4SRL both obtain zero-violation policies after convergence. But IC4SRL achieves much better reward performance than CPO (3x margin).

\section{Experiment Details}
\label{appendix: experiment details}
\subsection{Environment Settings}
\paragraph{Goal Task}
In the Goal task environments, the reward function is:
\begin{equation}\notag
\begin{split}
    & r(x_t) = d^{g}_{t-1} - d^{g}_{t} + \mathbbm{1}[d^g_t < R^g]~,\\
\end{split}
\end{equation}
where $d^g_t$ is the distance from the robot to its closest goal and $R^g$ is the size (radius) of the goal. When a goal is achieved, the goal location is randomly reset to someplace new while keeping the rest of the layout the same.

\paragraph{Push Task}
In the Push task environments, the reward function is:
\begin{equation}\notag
\begin{split}
    & r(x_t) = d^{r}_{t-1} - d^{r}_{t} + d^{b}_{t-1} - d^{b}_{t} + \mathbbm{1}[d^b_t < R^g]~,\\
\end{split}
\end{equation}
where $d^r$ and $d^b$ are the distance from the robot to its closest goal and the distance from the box to its closest goal, and $R^g$ is the size (radius) of the goal. The box size is $0.2$ for all the Push task environments. Like the goal task, a new goal location is drawn each time a goal is achieved. 


\paragraph{Hazard Constraint}
In the Hazard constraint environments, the cost function is:
\begin{equation}\notag
\begin{split}
    & c(x_t) = \max(0, R^h - d^h_t)~,\\
\end{split}
\end{equation}
where $d^h_t$ is the distance to the closest hazard and $R^h$ is the size (radius) of the hazard.
\paragraph{Pillar Constraint}
In the Pillar constraint environments, the cost $c_t = 1$ if the robot contacts with the pillar otherwise $c_t = 0$. 

\paragraph{State Space}
The state space is composed of various physical quantities from standard robot sensors (accelerometer, gyroscope, magnetometer, and velocimeter) and lidar (where each lidar sensor perceives objects of a single kind). The state spaces of all the test suites are summarized in \Cref{tab:state_space}. 
\begin{table*}[t]
\vskip 0.15in
\caption{The state space components of different test suites environments.}
\begin{center}
\begin{tabular}{c|cccc}
\toprule
\textbf{State Space Option} &  Goal-Hazard & Goal-Pillar & Push-Hazard & Push-Pillar\\
\hline
Accelerometer & \Checkmark & \Checkmark & \Checkmark  & \Checkmark \\
Gyroscope & \Checkmark & \Checkmark & \Checkmark  & \Checkmark \\
Magnetometer& \Checkmark & \Checkmark & \Checkmark  & \Checkmark\\
Velocimeter & \Checkmark & \Checkmark & \Checkmark  & \Checkmark\\
Goal Lidar & \Checkmark & \Checkmark & \Checkmark  & \Checkmark\\
Hazard Lidar & \Checkmark & \XSolid & \Checkmark & \XSolid\\
Pillar Lidar & \XSolid & \Checkmark & \XSolid  & \Checkmark\\
Box Lidar & \XSolid & \XSolid & \Checkmark & \Checkmark\\
\bottomrule
\end{tabular}
\label{tab:state_space}
\end{center}
\end{table*}

\subsection{Policy Settings}

Detailed parameter settings are shown in \Cref{tab:policy_setting}. All the policies in our experiments use the default hyper-parameter settings hand-tuned by Safety Gym~\cite{berkenkamp2017safe}. 
\begin{table*}[htbp]
\vskip 0.15in
\begin{center}
\begin{tabular}{c|ccc}
\toprule
\textbf{Policy Parameter} & PPO & PPO-Lagrangian \& PPO-Lagrangian+IC4SRL  & CPO \& CPO+IC4SRL \\
\hline
Timesteps per iteration & 30000 & 30000 & 30000\\
Policy network hidden layers & (256, 256) & (256, 256) & (256, 256) \\
Value network hidden layers & (256, 256) & (256, 256) & (256, 256) \\
Policy learning rate & 0.0004 & 0.0004 & (N/A)\\
Value learning rate & 0.001 & 0.001 & 0.001 \\
Target KL & 0.01 & 0.01 & 0.01 \\
Discounted factor $\gamma$ & 0.99 & 0.99 & 0.99  \\
Advantage discounted factor $\lambda$ & 0.97 & 0.97& 0.97  \\
PPO Clipping $\epsilon$ & 0.2 & 0.2  & (N/A)  \\
TRPO Conjugate gradient damping & (N/A) & (N/A) & 0.1  \\
TRPO Backtracking steps & (N/A) & (N/A) & 10 \\
Cost limit & (N/A) & 0 & 0 \\
\bottomrule
\end{tabular}
\caption{Important hyper-parameters of PPO, PPO-Lagrangian, CPO. Note that all the hyper-parameters are fixed the same across different environments with or without intrinsic cost functions.}
\label{tab:policy_setting}
\end{center}
\end{table*}

\subsection{Intrinsic Cost Settings}
\label{appendix: Intrinsic Cost Settings}
As introduced in \Cref{tab:state_space}, there are many components of the state space while some of the components are irrelevant to cost functions designs (e.g, lidar information of goal). Therefore, we only select cost-relevant information (i.e., lidar information of hazard/pillar) as the input features for the intrinsic cost function. 
The parameter space of the intrinsic cost function is the parameter space of the 1-hidden-layer (with four neurons) MLP we used. The input features of lidar information of constants are 8-dimensional. Together with the MLP (five hidden neurons and one output scale neuron), we have 41 parameters in total for the MLP of intrinsic cost functions including weights and bias.

\subsection{Computing Infrastructure}
\label{appendix: Computing Infrastructure}
We present the computing infrastructure and the corresponding computational time used in \Cref{tab:computing infrastructure} and \Cref{tab:computing time}. During the evolution, note that the training of each intrinsic cost candidate can be parallel. With sufficient computing power, the computing time of AutoCost only scales linearly with stages. Due to the limitation of our computing infrastructure, we can only make half of the training of one population (25 candidates / 50 candidates) parallel. The corresponding computing time is shown in \Cref{tab:computing time}.
\begin{table}[H]
\begin{center}
\begin{tabular}{c|c}
\toprule
CPU & AMD 3970X 32-Core Processor \\
\midrule
GPU & RTX2080TIx2 \\
\midrule
Memory & 256GB \\
\bottomrule
\end{tabular}
\caption{The computing infrastructure.}
\label{tab:computing infrastructure}
\end{center}
\end{table}

\begin{table}[H]
\begin{center}
\resizebox{\columnwidth}{!}{%
\begin{tabular}{c|c}
\toprule
Tasks & Computation time in hours \\
\midrule
Goal-Hazard-Point &  3.8 \\
Goal-Hazard-Pillar &  3.7 \\
Push-Hazard-Point &  4.4 \\
Push-Hazard-Pillar &  4.2 \\
Goal-Hazard-Car &  4.1 \\
Goal-Hazard-Doggo &  4.2 \\
\midrule
Goal-Hazard-Point (evoluation) &  54.1 \\
\bottomrule
\end{tabular}
}
\caption{The computing infrastructure.}
\label{tab:computing time}
\end{center}
\end{table}

\end{document}